\documentclass[12pt]{article}
\usepackage{tikz,amssymb,url,amsmath,amsthm,listings,algorithm,algpseudocode,enumerate, caption, subcaption,xspace, float, verbatim,tabularx,mathtools,adjustbox}
\usepackage[pointedenum]{paralist}
\usepackage{tikz}
\usetikzlibrary{decorations.pathreplacing}
\usepackage[authoryear]{natbib}
\usepackage{epstopdf}
\usepackage[breaklinks=true,pdfstartview=FitH]{hyperref}
\usepackage{times}
\usepackage{graphicx}
\usepackage{color}
\usepackage{multirow}
\usepackage{rotating}
\usepackage{bbm}
\usepackage{array}
\usepackage{latexsym}
\usepackage{longtable}
\usepackage{xr}
\usepackage{xcolor} 
\usepackage[symbol]{footmisc} 
\usepackage{dsfont}
\usepackage[numbered,framed]{matlab-prettifier}
\usepackage{blkarray, bigstrut}
\usepackage{xparse}
\usepackage{ragged2e}
\usetikzlibrary{fit}
\usetikzlibrary{positioning}
\let\oldnl\nl   
\newcommand{\nonl}{\renewcommand{\nl}{\let\nl\oldnl}}   

\newcounter{bar}

\makeatletter
\renewcommand*\env@matrix[1][*\c@MaxMatrixCols c]{%
    \hskip -\arraycolsep
    \let\@ifnextchar\new@ifnextchar
    \array{#1}}
\makeatother
\DeclarePairedDelimiter\floor{\lfloor}{\rfloor}

\def\bx{{\boldsymbol x}}
\def\by{{\boldsymbol y}}
\def\bv{{\boldsymbol v}}
\def\bb{{\boldsymbol b}}

\def\bd{{\boldsymbol d}}

\def\bp{{\boldsymbol p}}

\def\bs{{\boldsymbol s}}
\def\bu{{\boldsymbol u}}

\def\bq{{\boldsymbol q}}
\def\bz{{\boldsymbol z}}

\def\btheta{\boldsymbol \theta}

\def\arrayspace{\;\;}
\def\im2col{{\text{\tt im2col}}}
\def\accumarray{{\text{\tt accumarray}}}
\def\pad_idx{{\text{\tt pad\_idx}}}
\def\mean{{\text{mean}}}
\def\mat{\text{mat}}
\def\vec{{\text{vec}}}
\def\mfirst{m}
\def\msec{{m+1}}
\def\mout{{L+1}}
\def\inpu{{\text{\normalfont in}}}
\def\oupu{{\text{\normalfont out}}}
\def\padding{{\text{\normalfont pad}}}
\def\conv{{\text{\normalfont conv}}}
\def\pool{{\text{\normalfont pool}}}
\def\genphiZin{Z^{\inpu,i}}
\def\genphizin{z^{\inpu,i}}
\def\ain{{a^{\inpu}}}
\def\bin{{b^{\inpu}}}
\def\din{{d^{\inpu}}}
\def\aout{{a^{\oupu}}}
\def\bout{{b^{\oupu}}}

\def\genphis{{s}}
\def\genphih{{h}}
\def\padain{{a^{m}}}
\def\padbin{{b^{m}}}
\def\paddin{{d^{m}}}
\def\convain{{a^{m}_{\padding}}}
\def\convbin{{b^{m}_{\padding}}}
\def\convdin{{d^{m}}}
\def\convaout{{a^{m}_{\conv}}}
\def\convbout{{b^{m}_{\conv}}}
\def\convdout{{d^{m+1}}}
\def\poolain{{\convaout}}
\def\poolbin{{\convbout}}
\def\pooldin{{d^{m+1}}}
\def\poolaout{{a^{m+1}}}
\def\poolbout{{b^{m+1}}}
\def\pooldout{{d^{m+1}}}

\usepackage{tikz}
\usetikzlibrary{3d}
\makeatletter
\tikzoption{canvas is plane}[]{\@setOxy#1}
\def\@setOxy O(#1,#2,#3)x(#4,#5,#6)y(#7,#8,#9)%
  {\def\tikz@plane@origin{\pgfpointxyz{#1}{#2}{#3}}%
   \def\tikz@plane@x{\pgfpointxyz{#4}{#5}{#6}}%
   \def\tikz@plane@y{\pgfpointxyz{#7}{#8}{#9}}%
   \tikz@canvas@is@plane
  }
\makeatother

\renewcommand{\baselinestretch}{2}
\newcommand{\partialdiff}[2] {\frac{\partial #1}{\partial #2}}
\newcommand{\mpartialdiff}[2] {\frac{\partial #1}{\partial {#2}^{T}}}

\newcommand{\codedir}{anc}
\renewcommand{\thefootnote}{\normalsize \arabic{footnote}}
\allowdisplaybreaks

\algblock{ParFor}{EndParFor}
\algnewcommand\algorithmicparfor{\textbf{parfor}}
\algnewcommand\algorithmicpardo{\textbf{do}}
\algnewcommand\algorithmicendparfor{\textbf{end\ parfor}}
\algrenewtext{ParFor}[1]{\algorithmicparfor\ #1\ \algorithmicpardo}
\algrenewtext{EndParFor}{\algorithmicendparfor}

\begin{document}
\bigskip
\begin{center} {\LARGE Newton Methods for Convolutional Neural Networks}
\end{center}
\begin{center}
    {\bf \large Chien-Chih Wang, Kent Loong Tan, Chih-Jen Lin}\\
    {Department of Computer Science, National Taiwan University, Taipei
    10617, Taiwan}\\
\end{center}
\renewcommand{\thefootnote}{\fnsymbol{footnote}}
\renewcommand{\thefootnote}{\normalsize \arabic{footnote}}

\begin{abstract}
Deep learning involves a difficult non-convex optimization problem, which is often solved by stochastic gradient (SG) methods.
While SG is usually effective, it may not be robust in some situations.
Recently, Newton methods
have been investigated as an alternative optimization technique, but nearly all existing studies consider only fully-connected feedforward neural networks. 
They do not investigate other types of networks such as Convolutional Neural Networks (CNN), which are more commonly
used in deep-learning applications.
One reason is that Newton methods for CNN involve complicated operations, and so
far no works have conducted a thorough investigation. In this work, we give
details of all building blocks including function, gradient, and Jacobian evaluation,
and Gauss-Newton matrix-vector products. These basic components are very
important because with them further developments of Newton methods for CNN
become possible. We show that an efficient {\sl MATLAB} implementation can be
done in just several hundred lines of code and demonstrate that the Newton
method gives competitive test accuracy.
\end{abstract}

\markboth{}{NC instructions}
\ \vspace{-0mm}\\

\section{Introduction}
\label{sec:intro}
Deep learning is now widely used in many applications. To apply this technique, a difficult non-convex optimization problem must be solved.
Currently, stochastic gradient (SG) methods and their variants are the major optimization technique used for deep learning \citep[e.g., ][]{AK12b, KS14a}.
This situation is different from some application domains, where other types of optimization methods are more frequently used.
One interesting research question is thus to study if other optimization methods can be extended to be viable alternatives for deep learning.
In this work, we aim to address this issue by developing a practical Newton method for deep learning.

\par Some past works have studied Newton methods for training deep neural
networks (e.g., \citealt{JM10a, OV12a, RK13a, CCW15a, CCW16a, XH16a, AB17a}).
Almost all of them consider fully-connected feedforward neural networks and some have shown the potential of Newton methods for
being more robust than SG. Unfortunately, these works have not fully established Newton methods as a practical technique for deep learning
because other types of networks such as Convolutional Neural Networks (CNN) are more commonly used in deep-learning applications.
One important reason why CNN was not considered is because of the very complicated operations in implementing Newton methods.
Up to now no works have shown details of all the building blocks such as function, gradient, and Jacobian evaluation, and Hessian-vector products.
In particular, because interpreter-type languages such as {\sl Python} or {\sl MATLAB} have been popular for deep learning, how to easily implement
efficient operations by these languages is an important research issue.

\par In this work, we aim at a thorough investigation on the implementation of Newton methods for CNN. We focus on basic components
because without them none of any recent improvement of Newton methods for fully-connected networks can be even tried.
Our work will enable many further developments of Newton methods for CNN and maybe even other types of networks.

\par This paper is organized as follows. In Section \ref{sec:FunEval}, we
introduce CNN.
In Section \ref{sec:HessianFree}, Newton methods for CNN are investigated and
the detailed mathematical formulations of all operations are derived.
In Section \ref{sec:impl}, we provide details for an efficient 
{\sl MATLAB} implementation. The analysis of memory usage and computational
complexity is in Section \ref{sec:analysis}. Preliminary experiments to demonstrate the viability of Newton methods for CNN are in Section \ref{sec:exps}.
Section \ref{sec:conclu} concludes this work. A list of symbols is in the
appendix.

\par A simple and efficient {\sl MATLAB} implementation in just a few hundred
lines of code is available at
\begin{center}\url{https://www.csie.ntu.edu.tw/~cjlin/cnn/}\end{center}
Programs used for experiments in this paper and supplementary materials can be found at the same page.

\section{Optimization Problem of Convolutional Neural Networks}
\label{sec:FunEval}
Consider a $K$-class problem, where the training data set consists of $l$ input pairs $(Z^{1,i},$
$\by^i),\ i=1,\ldots,l$.
Here $Z^{1,i}$ is the $i$th input image with dimension $a^1
\times b^1 \times d^1$, where $a^1$ denotes the height of the input images, $b^1$ represents
the width of the input images and $d^1$ is the number of color channels.
If $Z^{1,i}$ belongs to the $k$th class, then the label vector is
\begin{equation*}
\by^i = [\underbrace{0,\ldots,0}_{k-1},1,0,\ldots,0]^T \in R^K.
\end{equation*}

A CNN \citep{YL89a} utilizes a stack of convolutional layers followed by fully-connected layers
to predict the target vector. 
Let $L^c$ be the number of convolutional layers, and $L^f$ be the number of
fully-connected layers. The number of layers is
\begin{equation*}
L = L^c + L^f.
\end{equation*}
Images
\begin{equation*}
Z^{1,i},\ i=1,\ldots,l
\end{equation*}
are input to the first layer, while the last (the $L$th) layer outputs a predicted label vector
\begin{equation*}
  \hat{\by}^{i},\ i=1,\ldots,l.
\end{equation*}
A hallmark of CNN is that both input and output of convolutional layers are
explicitly assumed to be images.

\subsection{Convolutional Layer}
\label{subsec:conv-layer}

In a convolutional layer, besides the main convolutional operations, two optional steps are padding and pooling,
each of which can also be considered as a layer with input/output images. To
easily describe all these operations in a unified setting, for the $i$th
instance, we assume the input image of the current layer is
\begin{equation*}
Z^{\inpu,i}
\end{equation*}
containing $d^{\inpu}$ channels of $a^{\inpu} \times b^{\inpu}$ images:
\begin{equation}
\label{Z-matrices}
\begin{bmatrix}
  z^{i}_{1,1,1} & & z^{i}_{1,b^{\inpu},1}\\
& \ddots & \\
z^{i}_{a^{\inpu},1,1} & & z^{i}_{a^{\inpu},b^{\inpu},1}
\end{bmatrix}
\quad
\hdots
\quad
\begin{bmatrix}
  z^{i}_{1,1,d^{\inpu}} & & z^{i}_{1,b^{\inpu},d^{\inpu}} \\
& \ddots & \\
  z^{i}_{a^{\inpu},1,d^{\inpu}} & & z^{i}_{a^{\inpu},b^{\inpu},d^{\inpu}}
\end{bmatrix}.
\end{equation}
The goal is to generate an output image
\begin{equation*}
Z^{\oupu,i}
\end{equation*}
of $d^{\oupu}$ channels of $a^{\oupu} \times b^{\oupu}$ images.

\par Now we describe details of convolutional operations.
To generate the output, we consider $d^{\oupu}$ filters, each of which is a $3$-D
weight matrix of size
\begin{equation*}
h \times h \times d^{\inpu}.
\end{equation*}
Specifically, the $j$th filter includes the following matrices
\begin{equation*}
\begin{bmatrix}
w^{j}_{1,1,1} & & w^{j}_{1,h,1} \\
& \ddots & \\
w^{j}_{h,1,1} & & w^{j}_{h,h,1}
\end{bmatrix},
  \quad
\hdots
\quad,
\begin{bmatrix}
w^{j}_{1,1,d^{\inpu}} & & w^{j}_{1,h,d^{\inpu}} \\
& \ddots & \\
w^{j}_{h,1,d^{\inpu}} & & w^{j}_{h,h,d^{\inpu}}
\end{bmatrix}
\end{equation*}
and a bias term $b_j$.

The main idea of CNN is to extract local information by convolutional operations, each of which is the
inner product between a small sub-image and a filter.
For the $j$th filter, we scan the entire input image to obtain small regions of
size $(h, h)$ and calculate the inner product between each region and the
filter. For example, if we start from the upper left corner of the input image,
  the first sub-image of channel $d$ is
\begin{equation*}
\begin{bmatrix}
z^{i}_{1,1,d} & \hdots & z^{i}_{1,h,d} \\
& \ddots & \\
z^{i}_{h,1,d} & \hdots & z^{i}_{h,h,d}
\end{bmatrix}.
\end{equation*}
We then calculate the following value.
\begin{equation}
\label{innerproduct1}
\sum^{d^{\inpu}}_{d=1}
\left<
\begin{bmatrix}
z^{i}_{1,1,d} & \hdots & z^{i}_{1,h,d} \\
& \ddots & \\
z^{i}_{h,1,d} & \hdots & z^{i}_{h,h,d}
\end{bmatrix},
\begin{bmatrix}
w^{j}_{1,1,d} & \hdots & w^{j}_{1,h,d} \\
& \ddots & \\
w^{j}_{h,1,d} & \hdots & w^{j}_{h,h,d}
\end{bmatrix}\right> + b_j,
\end{equation}
where $\left< \cdot, \cdot \right>$ means the sum of component-wise products between two
matrices. This value becomes the $(1, 1)$ position
of the channel $j$ of the output image.

\par Next, we must obtain other sub-images to produce values in other positions
of the output image. We specify the stride $s$ for sliding the filter. That
is, we move $s$ pixels vertically or horizontally to get sub-images. For the
$(2, 1)$ position of the output image, we move down $s$ pixels vertically to
obtain the following sub-image:
\begin{equation*}
\begin{bmatrix}
z^{i}_{1+s,1,d} & \hdots & z^{i}_{1+s,h,d} \\
& \ddots & \\
z^{i}_{h+s,1,d} & \hdots & z^{i}_{h+s,h,d}
\end{bmatrix}.
\end{equation*}
Then the $(2,1)$ position of the channel $j$ of the output image is
\begin{equation}
\label{innerproduct2}
\sum^{d^{\inpu}}_{d=1}
\left<
\begin{bmatrix}
z^{i}_{1+s,1,d} & \hdots & z^{i}_{1+s,h,d} \\
& \ddots & \\
z^{i}_{h+s,1,d} & \hdots & z^{i}_{h+s,h,d}
\end{bmatrix},
\begin{bmatrix}
w^{j}_{1,1,d} & \hdots & w^{j}_{1,h,d} \\
& \ddots & \\
w^{j}_{h,1,d} & \hdots & w^{j}_{h,h,d}
\end{bmatrix}\right> + b_j.
\end{equation}
Assume that vertically and horizontally we can move the filter $a^{\oupu}$ and $b^{\oupu}$
times, respectively. Therefore,
\begin{align}
\label{abs}
a^{\oupu} &= \floor{\frac{a^{\inpu} - h}{s}} + 1, \nonumber\\
b^{\oupu} &= \floor{\frac{b^{\inpu} - h}{s}} + 1.
\end{align}

For efficient implementations, we can conduct all operations including
\eqref{innerproduct1}
and \eqref{innerproduct2} by matrix operations. To begin,
we concatenate the matrices of the different channels in \eqref{Z-matrices} to
\begin{equation}
\label{zm-1}
Z^{\inpu,i} =
\begin{bmatrix}
z^{i}_{1,1,1} & \hdots & z^{i}_{a^{\inpu},1,1} &
  z^{i}_{1,2,1} & \hdots & z^{i}_{a^{\inpu},b^{\inpu},1} \\
\vdots & \ddots & \vdots & \vdots & \ddots & \vdots \\
z^{i}_{1,1,d^{\inpu}} & \hdots &
  z^{i}_{a^{\inpu},1,d^{\inpu}} & z^{i}_{1,2,d^{\inpu}} &
  \hdots & z^{i}_{a^{\inpu},b^{\inpu},d^{\inpu}}
\end{bmatrix},\ i=1,\ldots,l.
\end{equation}
We note that
\eqref{innerproduct1} is the inner product between the following two vectors
\begin{equation*}
\begin{bmatrix}
w^{j}_{1,1,1} & \hdots & w^{j}_{h,1,1} & w^{j}_{1,2,1} & \hdots &
  w^{j}_{h,h,1} & \hdots & w^{j}_{h, h, d^{\inpu}} & b_j
\end{bmatrix}^T
\end{equation*}
and
\begin{equation*}
\begin{bmatrix}
z^{i}_{1,1,1} & \hdots & z^{i}_{h, 1, 1} & z^{i}_{1,
2, 1} & \hdots & z^{i}_{h, h, 1} & \hdots & z^{i}_{h, h,
d^{\inpu}}& 1
\end{bmatrix}^T.
\end{equation*}
Based on \cite{AV15a}, we define the following two operators
\begin{align}
  \vec(M) &= \begin{bmatrix}
  M_{:,1}  \\
\vdots  \\
M_{:,b}
\end{bmatrix} \in R^{ab \times 1}, \text{ where } M \in R^{a \times b}, \label{matrix-vector}\\
\mat(\bv)_{a \times b} &= 
\begin{bmatrix}  
v_1 & & v_{(b-1)a + 1} \\
\vdots & \cdots & \vdots\\
v_a & & v_{ba}
\end{bmatrix}\in R^{a \times b}, \text{ where } \bv \in R^{ab \times 1}.
\label{vector-matrix}
\end{align}
There exists a $0/1$ matrix
\begin{equation*}
  P_{\phi} \in R^{h h d^{\inpu} a^{\oupu} b^{\oupu} 
  \times d^{\inpu} a^{\inpu} b^{\inpu}}
\end{equation*}
so that an operator
\begin{align*}
  \phi:&\ R^{d^{\inpu}\times a^{\inpu}b^{\inpu}} \rightarrow
  R^{h h d^{\inpu} \times a^{\oupu} b^{\oupu}}
\end{align*}
defined as
\begin{equation}
\label{fun-phi}
	\phi(Z^{\inpu,i}) \equiv \mat\left(P_{\phi}
  \vec(Z^{\inpu,i})\right)_{h h d^{\inpu}
  \times a^{\oupu} b^{\oupu}},\ \forall i,
\end{equation}
collects all sub-images in $Z^{\inpu,i}$.
Specifically, $\phi(Z^{\inpu,i})$ is
\begin{equation}
\label{big-zm-1}
\begin{bmatrix}
  z^{i}_{1,1,1} & \hdots & z^{i}_{1+(a^{\oupu}-1)\times s,1,1}
  & z^{i}_{1,1+s,1} & \hdots & z^{i}_{1 + (a^{\oupu}-1) \times
s, 1 + (b^{\oupu}-1) \times s, 1} \\
  z^{i}_{2,1,1} & \hdots & z^{i}_{2+(a^{\oupu}-1)\times s,1,1}
& z^{i}_{2,1+s,1} & \hdots & z^{i}_{2 + (a^{\oupu}-1) \times
s, 1 + (b^{\oupu}-1) \times s, 1} \\
\vdots & \ddots & \vdots & \vdots & \ddots & \vdots \\
  z^{i}_{h,h,1} & \hdots & z^{i}_{h+(a^{\oupu}-1)\times
  s,h,1} & z^{i}_{h,h+s,1} & \hdots & z^{i}_{h +
  (a^{\oupu}-1) \times s, h + (b^{\oupu}-1) \times s, 1} \\
\vdots & \ddots & \vdots & \vdots & \ddots & \vdots \\
  z^{i}_{1,1,d^{\inpu}} & \hdots &
  z^{i}_{1+(a^{\oupu}-1)\times s,1,d^{\inpu}} &
  z^{i}_{1,1+s,d^{\inpu}} & \hdots & z^{i}_{1 +
(a^{\oupu}-1) \times s, 1 + (b^{\oupu}-1) \times s, d^{\inpu}} \\
\vdots & \ddots & \vdots & \vdots & \ddots & \vdots \\
  z^{i}_{h,h,d^{\inpu}} & \hdots &
z^{i}_{h+(a^{\oupu}-1)\times s,h,d^{\inpu}} &
z^{i}_{h,h+s,d^{\inpu}} & \hdots & z^{i}_{h +
(a^{\oupu}-1) \times s, h + (b^{\oupu}-1) \times s, d^{\inpu}}
\end{bmatrix}.
\end{equation}
By considering
\begin{equation}
\label{conv-wmatrix}
W = 
\begin{bmatrix}
w^{1}_{1,1,1} & w^{1}_{2,1,1} & \hdots & w^{1}_{h, h, d^{\inpu}}\\
  \vdots  & \vdots & \ddots & \vdots \\
  w^{d^{\oupu}}_{1,1,1} & w^{d^{\oupu}}_{2,1,1} & \hdots & w^{d^{\oupu}}_{h,h, d^{\inpu}}
\end{bmatrix}\in R^{d^{\oupu} \times h h d^{\inpu}} \text{ and }
\bb = 
\begin{bmatrix}
  b_1 \\ \vdots \\ b_{d^{\oupu}}
\end{bmatrix} \in R^{d^{\oupu} \times 1},
\end{equation}
all convolutional operations can be combined as
\begin{equation}
\label{conv-f-ztos}
S^{\oupu,i} = W \phi(Z^{\inpu,i}) + \bb \mathds{1}^T_{a^{\oupu}
b^{\oupu}} \in R^{d^{\oupu} \times a^{\oupu} b^{\oupu}},
\end{equation}
where
\begin{equation*}
S^{\oupu,i} =
\begin{bmatrix}
  s^{i}_{1,1,1} & \hdots & s^{i}_{a^{\oupu},1,1} & s^{i}_{1,2,1} & \hdots
  & s^{i}_{a^{\oupu},b^{\oupu},1} \\
\vdots & \ddots & \vdots & \vdots & \ddots & \vdots \\
  s^{i}_{1,1,d^{\oupu}} & \hdots & s^{i}_{a^{\oupu},1,d^{\oupu}}
  & s^{i}_{1,2,d^{\oupu}} & \hdots & s^{i}_{a^{\oupu},b^{\oupu},d^{\oupu}}
\end{bmatrix}
\end{equation*}
  and
\begin{equation*}
\mathds{1}_{a^{\oupu} b^{\oupu}}=
\begin{bmatrix} 1 \\ \vdots \\ 1 \end{bmatrix} \in R^{a^{\oupu} b^{\oupu} \times 1}.
\end{equation*}

Next, an activation function scales each element of $S^{\oupu,i}$ to obtain the
output matrix $Z^{\oupu,i}$.
\begin{equation}
\label{conv-f-stoz}
Z^{\oupu,i} = \sigma(S^{\oupu,i}) \in R^{d^{\oupu} \times a^{\oupu} b^{\oupu}}.
\end{equation}
For CNN, commonly the following RELU activation function
\begin{equation}
\label{relu}
\sigma(x) = \max(x,0)
\end{equation}
is used and we consider it in our implementation.\footnote{To use Newton methods,
$\sigma(x)$ should be twice differentiable, but the RELU function is not. For
simplicity, we follow \cite{AK12b} to assume $\sigma'(x) = 1$ if $x>0$ and $0$
otherwise. It is possible to use a differentiable approximation of the RELU
function, though we leave this issue for future investigation.}

\par Note that by the matrix representation, the storage is increased from
\begin{equation*}
 d^{\inpu} \times a^{\inpu} b^{\inpu} 
\end{equation*}
in \eqref{Z-matrices} to
\begin{equation*}
  (h h d^{\inpu}) \times a^{\oupu} b^{\oupu}
\end{equation*}
in \eqref{big-zm-1}.
From \eqref{abs}, roughly
\begin{equation*}
\left(\frac{h}{s}\right)^2
\end{equation*}
folds increase of the memory occurs. However, we gain efficiency by using fast matrix-matrix
multiplications in optimized BLAS \citep{JJD90a}.

\subsubsection{Zero-padding}
\label{subsec:zero-padding}

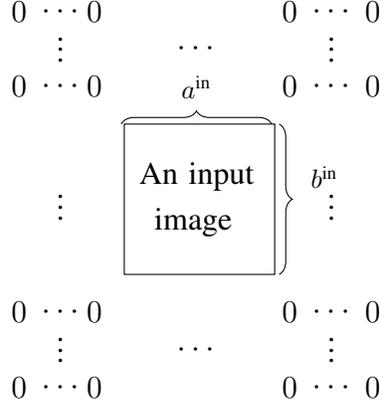
\begin{figure}[t]
\begin{center}
\begin{tikzpicture}
\draw (0,0) -- (2,0) -- (2,2) -- (0,2) -- (0,0);
\node [text width=3cm] at (1.7,1.3) {An input};
\node [text width=3cm] at (1.9,0.7) {image};
\node [text width=0cm] at (2.1,3.5) {$0$};
\node [text width=0cm] at (2.5,3.5) {$\cdots$};
\node [text width=0cm] at (3.2,3.5) {$0$};
\node [text width=0cm] at (2.7,3.1) {$\vdots$};
\node [text width=0cm] at (2.1,2.5) {$0$};
\node [text width=0cm] at (2.5,2.5) {$\cdots$};
\node [text width=0cm] at (3.2,2.5) {$0$};
\node [text width=0cm] at (2.7,1) {$\vdots$};
\node [text width=0cm] at (-0.9,1) {$\vdots$};
\node [text width=0cm] at (0.7,3) {$\cdots$};
\node [text width=0cm] at (0.7,-1) {$\cdots$};
\node [text width=0cm] at (2.1,-0.5) {$0$};
\node [text width=0cm] at (2.5,-0.5) {$\cdots$};
\node [text width=0cm] at (3.2,-0.5) {$0$};
\node [text width=0cm] at (2.7,-0.9) {$\vdots$};
\node [text width=0cm] at (2.1,-1.5) {$0$};
\node [text width=0cm] at (2.5,-1.5) {$\cdots$};
\node [text width=0cm] at (3.2,-1.5) {$0$};
\node [text width=0cm] at (-0.5,-0.5) {$0$};
\node [text width=0cm] at (-1.1,-0.5) {$\cdots$};
\node [text width=0cm] at (-1.5,-0.5) {$0$};
\node [text width=0cm] at (-0.9,-0.9) {$\vdots$};
\node [text width=0cm] at (-0.5,-1.5) {$0$};
\node [text width=0cm] at (-1.1,-1.5) {$\cdots$};
\node [text width=0cm] at (-1.5,-1.5) {$0$};
\node [text width=0cm] at (-0.5,3.5) {$0$};
\node [text width=0cm] at (-1.1,3.5) {$\cdots$};
\node [text width=0cm] at (-1.5,3.5) {$0$};
\node [text width=0cm] at (-0.9,3.1) {$\vdots$};
\node [text width=0cm] at (-0.5,2.5) {$0$};
\node [text width=0cm] at (-1.1,2.5) {$\cdots$};
\node [text width=0cm] at (-1.5,2.5) {$0$};
\draw [decorate,decoration={brace,amplitude=5pt},xshift=-4pt,yshift=0pt] (0.1,2)
-- (2.1,2) node [above,midway,yshift=0.2cm] {\footnotesize $a^{\inpu}$};
\draw [decorate,decoration={brace,amplitude=5pt},xshift=-4pt,yshift=0pt] (2.2,2)
-- (2.2,0) node [above,midway,xshift=0.6cm] {\footnotesize $b^{\inpu}$};
\end{tikzpicture}
\end{center}
\caption{An illustration of the padding operation to have zeros around the
border.}
\label{fig:padding}
\end{figure}

To better control the size of the output image, before the convolutional
operation we may enlarge the input image to have zero values around the border.
This technique is called zero-padding in CNN training. 
See an illustration in Figure \ref{fig:padding}. 

To specify the mathematical operation we can treat the padding operation as a layer of mapping
an input $Z^{\inpu,i}$ to an output $Z^{\oupu,i}$. Let
\begin{equation*}
d^{\oupu} = d^{\inpu}.
\end{equation*}
There exists a $0/1$ matrix
\begin{equation*}
  P_{\padding} \in R^{d^{\oupu} a^{\oupu} b^{\oupu} \times
  d^{\inpu} a^{\inpu} b^{\inpu}}
\end{equation*}
so that the padding operation can be represented by
\begin{equation}
  Z^{\oupu,i} \equiv \mat(P_{\padding}
  \vec(Z^{\inpu,i}))_{d^{\oupu} \times a^{\oupu} b^{\oupu}}.
  \label{padding}
\end{equation}

\subsubsection{Pooling Operations}
\label{subsubsec:pooling-layer}

\begin{figure}[t]
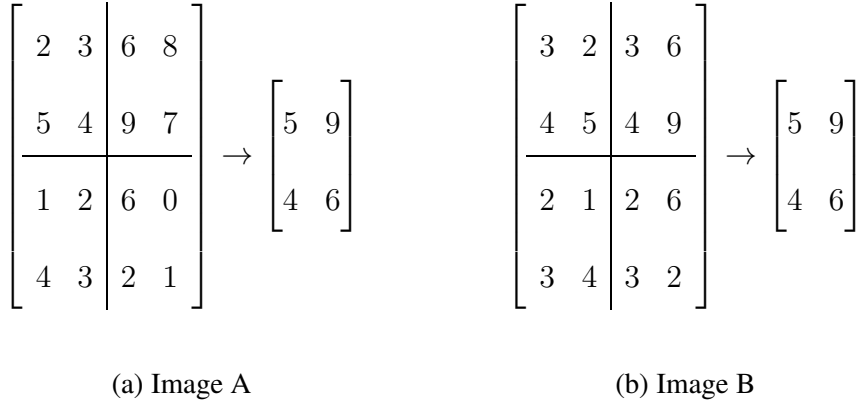

\begin{center}
	\begin{subfigure}{\textwidth}
		\begin{tabular}{c c}
		\begin{subfigure}{0.45\textwidth}
		\begin{equation*}
		\left[
		\begin{array}{cc|cc}
		2 & 3 & 6 & 8 \\
		5 & 4 & 9 & 7 \\
		\hline
		1 & 2 & 6 & 0 \\
		4 & 3 & 2 & 1
		\end{array}
		\right]
		\rightarrow
		\begin{bmatrix}
		5 & 9 \\
		4 & 6
		\end{bmatrix}
		\end{equation*}
		\caption{Image A}
		\end{subfigure}
		\begin{subfigure}{0.45\textwidth}
		\begin{equation*}
		\left[
		\begin{array}{cc|cc}
		3 & 2 & 3 & 6 \\
		4 & 5 & 4 & 9 \\
		\hline
		2 & 1 & 2 & 6 \\
		3 & 4 & 3 & 2
		\end{array}
		\right]
		\rightarrow
		\begin{bmatrix}
		5 & 9 \\
		4 & 6
		\end{bmatrix}
		\end{equation*}
		\caption{Image B}
		\end{subfigure}
		\end{tabular}
	\end{subfigure}
\end{center}
\caption{An illustration of max pooling to extract translational invariance features. 
The image B is derived from shifting A by $1$ pixel in the horizontal direction.}
\label{pooling-example}
\end{figure}

For CNN, to reduce the computational cost, a dimension reduction is often
applied by a pooling step after convolutional operations.
Usually we consider an operation that can (approximately) extract rotational or translational
invariance features.
Among the various types of pooling methods such as average pooling, max pooling,
and stochastic pooling,
we consider max pooling as an illustration because it is the most used setting for CNN.
We show an example of max pooling by considering two $4 \times 4$ images, A and B, in Figure \ref{pooling-example}.
The image B is derived by shifting A by $1$ pixel in the horizontal direction.
We split two images into four $2 \times 2$ sub-images and choose the max value from every sub-image. 
In each sub-image because only some elements are changed, the maximal value is likely the same or similar. This is called translational
invariance and for our example the two output images from A and B are the same.

Now we derive the mathematical representation.
Similar to Section \ref{subsec:zero-padding}, we consider the operation as a
separate layer for the easy description though in our implementation pooling is just an
operation at the end of the convolutional layer. Assume $Z^{\inpu,i}$ is an
input image. We partition every channel of $Z^{\inpu,i}$ into non-overlapping
sub-regions by $h \times h$ filters with the stride $s = h$.\footnote{
Because of the disjoint sub-regions, the stride $s$ for sliding the filters
is equal to $h$.}
This partition step is a special case of how we generate sub-images in convolutional operations.
Therefore, by the same definition as \eqref{fun-phi} we can generate the matrix
\begin{equation}
  \phi(Z^{\inpu,i}) = \mat(P_{\phi}\vec(Z^{\inpu,i}))_{h h \times
    d^{\oupu} a^{\oupu} b^{\oupu}},
\label{pool:phi}
\end{equation}
where
\begin{equation}
  a^{\oupu} = \floor{\frac{a^{\inpu}}{h}},\ b^{\oupu} =
  \floor{\frac{b^{\inpu}}{h}},\ d^{\oupu} = d^{\inpu}.
\label{pool:output-a-b}
\end{equation}
To select the largest element of each sub-region, there exists a matrix
\begin{equation*}
  W^{i} \in R^{d^{\oupu} a^{\oupu} b^{\oupu} \times h h d^{\oupu}
  a^{\oupu} b^{\oupu}}
\end{equation*}
so that each row of $W^{i}$ selects a single element from
$\vec(\phi(Z^{\inpu,i}))$.
Therefore,
\begin{equation}
  Z^{\oupu,i} = \mat\left(W^{i} \vec(\phi(Z^{\inpu,i}))\right)_{d^{\oupu} \times
  a^{\oupu} b^{\oupu}}.\label{maxpooling}
\end{equation}
Note that different from \eqref{conv-f-ztos} of the convolutional layer, $W^{i}$ is a constant matrix rather than a weight matrix.

By combining \eqref{pool:phi} and \eqref{maxpooling}, we have
\begin{equation}
  Z^{\oupu,i} = \mat\left(P^{i}_{\pool}\vec(Z^{\inpu,i})\right)_{d^{\oupu} \times
  a^{\oupu} b^{\oupu}},
\label{simple-maxpooling}
\end{equation}
where
\begin{equation}
  P^{i}_{\pool} = W^{i} P_{\phi} \in R^{d^{\oupu} a^{\oupu} b^{\oupu}
  \times d^{\inpu} a^{\inpu} b^{\inpu}}.
\label{P_pool}
\end{equation}

\subsubsection{Summary of a Convolutional Layer}
\label{subsubsec:summary-convlayer}

For the practical implementation, we find it is more suitable to consider padding and pooling 
as part of the convolutional layers. Here we discuss details of considering all operations together.
The whole convolutional layer involves the following procedure:
\begin{align}
Z^{m,i} &\rightarrow \text{padding by } \eqref{padding} \rightarrow \text{convolutional operations by } \eqref{conv-f-ztos}, \eqref{conv-f-stoz} \nonumber \\
      &\rightarrow \text{pooling by } \eqref{simple-maxpooling} \rightarrow
  Z^{m+1,i},
\label{flow-conv}
\end{align}
where $Z^{m,i}$ and $Z^{m+1,i}$ are input and output of the $m$th layer, respectively.

We use the following symbols to denote image sizes at different stages of the convolutional layer.
\begin{align*}
&a^m,\ b^m: \text{ size in the beginning} \\
&a^m_{\padding},\ b^m_{\padding}: \text{ size after padding} \\
&a^m_{\conv},\ b^m_{\conv}: \text{ size after convolution.}
\end{align*}
Table \ref{operation-table} indicates how these values are
$a^{\inpu},b^{\inpu},d^{\inpu}$ and $a^{\oupu},b^{\oupu},d^{\oupu}$ at different stages.
\begin{table}[t]
  \caption{Detailed information of operations at a convolutional layer.}
  \label{operation-table}
\begin{center}
\begin{tabular}{m{3.5cm} m{3.3cm} m{3.3cm} l l}
Operation  & $a^{\inpu},\ b^{\inpu},\ d^{\inpu}$ & $a^{\oupu},\ b^{\oupu},\
  d^{\oupu}$ & Input & Output\\
\hline
\text{Padding: } \eqref{padding} & $a^{m},\ b^m,\ d^m$ & $a^m_{\padding},\
b^m_{\padding},\ d^m$ & $Z^{m,i}$ & $\padding(Z^{m,i})$\\
\text{Convolution: } \eqref{conv-f-ztos} & $a^m_{\padding},\ b^m_{\padding},\
d^m$ & $a^m_{\conv},\ b^m_{\conv},\ d^{m+1}$ & $\padding(Z^{m,i})$ & $S^{m,i}$\\
\text{Convolution: } \eqref{conv-f-stoz} & $a^m_{\conv},\ b^m_{\conv},\ d^{m+1}$
& $a^m_{\conv},\ b^m_{\conv},\ d^{m+1}$ & $S^{m,i}$ & $\sigma(S^{m,i})$\\
\text{Pooling: } \eqref{simple-maxpooling} & $a^m_{\conv},\ b^m_{\conv},\
d^{m+1}$ & $a^{m+1},\ b^{m+1},\ d^{m+1}$ & $\sigma(S^{m,i})$ & $Z^{m+1,i}$\\
\end{tabular}
\end{center}
\end{table}
We further denote the filter size, mapping matrices and weight
matrices at the $m$th layer as
\begin{equation*}
h^m,\ P^m_{\padding},\ P^{m}_{\phi},\ P^{m,i}_{\pool},\ W^m,\ \bb^m.
\end{equation*}
Then from \eqref{padding}, \eqref{conv-f-ztos}, \eqref{conv-f-stoz},
\eqref{simple-maxpooling}, and Table \ref{operation-table}, all operations can
be summarized as
\begin{equation}
	Z^{m+1,i} = \mat(P^{m,i}_{\pool}\vec(\sigma(S^{m,i})))_{d^{m+1} \times a^{m+1}b^{m+1}},
\label{pool-Z}
\end{equation}
where
\begin{equation*}
S^{m,i} = W^m\mat(P^m_{\phi} P^m_{\padding} \vec(Z^{m,i}))_{h^m h^m d^m \times a^m_{\conv} b^m_{\conv}} + \bb^m.
\end{equation*}

\subsection{Fully-Connected Layer}
\label{subsec:full-layer}
After passing through the convolutional layers, we concatenate
columns in the matrix in \eqref{pool-Z} to form the input vector of the first
fully-connected layer.
\begin{equation*}
\bz^{\mfirst,i} = \vec(Z^{\mfirst,i}),\ i = 1, \ldots, l,\ m = L^c + 1.
\end{equation*}

\par In the fully-connected layers $(L^c < m \leq L)$, we consider the following
weight matrix and bias vector between layers $\mfirst$ and $\msec$.
\begin{equation}
W^{m} =
\begin{bmatrix}
w^m_{11}& w^m_{21}& \cdots & w^m_{n_{\mfirst}1}\\
w^m_{12}& w^m_{22}& \cdots & w^m_{n_{\mfirst}2}\\
\vdots& \vdots& \vdots & \vdots\\
w^m_{1n_{\msec}}& w^m_{2n_{\msec}}& \cdots & w^m_{n_{\msec}n_{\mfirst}}
\end{bmatrix}_{ n_{\msec} \times n_{\mfirst}}
\text{ and }\ 
\bb^m = 
\begin{bmatrix}
  b^m_1 \\ b^m_2 \\ \vdots \\ b^m_{n_{\msec}}
\end{bmatrix}_{n_{\msec} \times 1},
\label{fc-wmatrix}
\end{equation}
where $n_{\mfirst}$ and $n_{\msec}$ are the numbers of neurons in layers
$\mfirst$ and $\msec$, respectively.\footnote{
$n_{L^c+1} = d^{L^c+1} a^{L^c+1} b^{L^c+1}$ and $n_\mout = K$ is the number of classes.}
If $\bz^{\mfirst,i} \in R^{n_{\mfirst}}$ is the input vector, the following
operations are applied to generate the output vector $\bz^{\msec,i} \in
R^{n_{\msec}}$.
\begin{align}
  \bs^{\mfirst,i} &= W^m \bz^{\mfirst,i} + \bb^m, \label{full-f-s}\\
  z^{\msec,i}_j &= \sigma(s^{\mfirst,i}_j),\ j=1,\ldots,n_{\msec} \label{full-f-stoz}.
\end{align}
For the activation function in fully-connected layers, except the last layer, we also consider the RELU function defined in \eqref{relu}.
For the last layer, we use the following linear function.
\begin{equation}
\label{linear}
\sigma(x) = x.
\end{equation}

\subsection{The Overall Optimization Problem}
\label{subsec:overall-prob}
At the last layer, the output $\bz^{\mout,i}, \forall i$ is obtained.
We can apply a loss function to check how close $\bz^{\mout,i}$ is to the label vector
$\by^i$. In this work the following squared loss is considered.
\begin{equation}
\label{square-loss}
	\xi(\bz^{\mout,i}; \by^i) = ||\bz^{\mout,i} - \by^{i}||^2.
\end{equation}

We can collect all model parameters such as filters of convolutional layers in \eqref{conv-wmatrix} and
weights/biases in \eqref{fc-wmatrix} for fully-connected layers into a long vector $\btheta \in R^n$, where
$n$ becomes the total number of variables from the discussion in this section. 
\begin{equation*}
  n = \sum_{m=1}^{L^c} d^{\msec} \times (h^m \times h^m \times d^{\mfirst} + 1)
  + \sum_{m = L^c + 1}^L n_{\msec} \times (n_{\mfirst} + 1).
\end{equation*}

The output $\bz^{\mout,i}$ of the last layer is a function of $\btheta$.
The optimization problem to train a CNN is
\begin{equation}
\label{opt-problem}
      \min_{\btheta} f(\btheta),
\end{equation}
where
\begin{equation}
\label{L2loss}
f(\btheta) = \frac{1}{2C} \btheta^T \btheta + \frac{1}{l}\sum_{i=1}^l
\xi(\bz^{\mout,i}; \by^i).
\end{equation}
In \eqref{L2loss}, the second term is the average training loss, while a
regularization term with the parameter $C > 0$ is used to avoid overfitting.

\section{Hessian-free Newton Methods for Training CNN}
\label{sec:HessianFree}
To solve an unconstrained minimization problem such as \eqref{opt-problem},
a Newton method iteratively finds a search direction $\bd$ by solving the following second-order approximation.
\begin{equation}
\label{second-order-f}
\min_{\bd} \, \nabla f(\btheta)^T \bd + \frac{1}{2} \bd^T \nabla^2 f(\btheta) \bd,
\end{equation}
where $\nabla f(\btheta)$ and $\nabla^2 f(\btheta)$ are the gradient vector and the Hessian matrix, respectively.
In this section we present details of applying a Newton method to solve the CNN
problem \eqref{opt-problem}.

\subsection{Procedure of the Newton Method}
\label{subsec:proposed-alg}
For CNN, the gradient of $f(\btheta)$ is
\begin{equation}
\label{whole-gradient}
    \nabla f(\btheta) =  \frac{1}{C}{\btheta} + \frac{1}{l} \sum_{i=1}^l (J^i)^T
    \nabla_{\bz^{\mout,i}} \xi(\bz^{\mout,i};\by^i),
\end{equation}
where
\begin{equation}
\label{jacobian}
J^i =
\begin{bmatrix}
\frac{\partial z_1^{\mout,i}}{\partial \theta_1} & \cdots & \frac{\partial
z_1^{\mout,i}}{\partial \theta_n} \\
\vdots & \vdots & \vdots\\
\frac{\partial z_{n_\mout}^{\mout,i}}{\partial \theta_1} & \cdots & \frac{\partial
z_{n_\mout}^{\mout,i}}{\partial \theta_n}
\end{bmatrix}_{n_{\mout} \times n},\ i=1,\ldots,l,
\end{equation}
is the Jacobian of $\bz^{L+1,i}$.
The Hessian matrix of $f(\btheta)$ is
\begin{align}
\label{hessiantogauss}
\nabla^2 f(\btheta) =&  \frac{1}{C} \mathcal I + \frac{1}{l} \sum_{i=1}^l 
(J^i)^T B^{i}J^i
\nonumber\\
&+\frac{1}{l} \sum_{i=1}^l\sum_{j=1}^{n_{\mout}} \frac{\partial
\xi(\bz^{\mout,i};\by^i)}{\partial z_j^{\mout,i}}
\begin{bmatrix}
\frac{\partial^2 z_j^{\mout,i}}{\partial \theta_1 \partial \theta_1} & \cdots &
  \frac{\partial^2 z_j^{\mout,i}}{\partial \theta_1 \partial \theta_n}\\
\vdots & \ddots & \vdots\\
\frac{\partial^2 z_j^{\mout,i}}{\partial \theta_n \partial \theta_1} & \cdots &
  \frac{\partial^2 z_j^{\mout,i}}{\partial \theta_n \partial \theta_n}
\end{bmatrix},
\end{align}
where $\mathcal I$ is the identity matrix and
\begin{equation}
    B^{i}_{ts} = \frac{\partial^2 \xi(\bz^{\mout,i};\by^i)}{\partial z_t^{\mout,i}
    \partial z_s^{\mout,i}},\ t=1,\ldots,n_{\mout},\ s=1,\ldots,n_{\mout}.
\label{Bts}
\end{equation}
From now on for simplicity we let
\begin{equation*}
\xi_i \equiv \xi_i(\bz^{\mout,i};\by^i).
\end{equation*}

\par In general \eqref{hessiantogauss} is not positive semi-definite, so $f(\btheta)$ is
non-convex for deep learning. The sub-problem \eqref{second-order-f} is
difficult to solve and the resulting direction may not lead to the decrease of
the function value.
Following past works \citep{JM10a, CCW15a}, we consider the following
Gauss-Newton approximation \citep{NNS02a}
\begin{equation}
\label{GapproxH}
G = \frac{1}{C} \mathcal I + \frac{1}{l} \sum_{i=1}^l (J^i)^T B^{i}J^i \approx
\nabla^2 f(\btheta).
\end{equation}
In particular, if $G$ is positive definite, then \eqref{second-order-f} becomes the same as solving the following linear system.
\begin{equation}
\label{cgobj}
G\bd = -\nabla f(\btheta).
\end{equation}

\par After a Newton direction $\bd$ is obtained, to ensure the convergence, 
we update $\btheta$ by
\begin{equation*}
\btheta \leftarrow \btheta + \alpha \bd,
\end{equation*}
where $\alpha$ is the largest element in an exponentially decreased sequence
like $\{1, \frac{1}{2}, \frac{1}{4}, \ldots\}$ satisfying
\begin{equation}
\label{linesearch-condition}
f(\btheta + \alpha \bd) \leq f(\btheta) + \eta \alpha \nabla f(\btheta)^T \bd.
\end{equation}
In \eqref{linesearch-condition}, $\eta \in (0,1)$ is a pre-defined constant.
The procedure to find $\alpha$ is called a backtracking line search.

Past works \citep[e.g., ][]{JM10a, CCW16a} have shown that sometimes \eqref{cgobj} is too aggressive, so a direction closer to the negative gradient is better.
To this end, in recent works of applying Newton methods on fully-connected
networks, the Levenberg-Marquardt method \citep{KL44a,DWM63a} is used to solve
the following linear system rather than \eqref{cgobj}.
\begin{equation}
\label{cgobj-LM}
(G + \lambda \mathcal I)\bd = -\nabla f(\btheta),
\end{equation}
where $\lambda$ is a parameter decided by how good the function reduction is.
Specifically, if $\btheta + \bd$ is the next iterate after line search, we define
\begin{equation*}
  \rho = \frac{f(\btheta + \bd) - f(\btheta)}{ \nabla f(\btheta)^T \bd + \frac{1}{2} (\bd)^T G \bd}
\end{equation*}
as the ratio between the actual function reduction and the predicted reduction. By using $\rho$, the parameter $\lambda_{\text{next}}$ for the next iteration is decided by 
\begin{equation}
\label{LM-rules}
\lambda_{\text{next}} =
\begin{cases}
\lambda \times \text{drop}& \rho > \rho_{\text{upper}}, \\
\lambda & \rho_{\text{lower}} \leq \rho \leq \rho_{\text{upper}}, \\
\lambda \times \text{boost}& \text{otherwise,}
\end{cases}
\end{equation}
where (drop,boost) are given constants.
From \eqref{LM-rules} we can clearly see that if the function-value reduction is not satisfactory, then $\lambda$ is enlarged and the resulting direction
is closer to the negative gradient.

Next, we discuss how to solve the linear system \eqref{cgobj-LM}.
When the number of variables $n$ is large, the matrix $G$ is too large to be stored. For some optimization problems including neural networks, without explicitly storing $G$
it is possible to calculate the product between $G$ and any vector $\bv$ \citep{JM10a,JN11a,CCW16a}.
For example, from \eqref{GapproxH}, 
\begin{equation}
(G + \lambda \mathcal{I})\bv = (\frac{1}{C} + \lambda)\bv + \frac{1}{l} \sum_{i=1}^l \left((J^i)^T \left(B^i (J^i \bv)\right)\right).
\label{GV}
\end{equation}
If the product between $J^i$ and a vector can be easily calculated, then $G$ does not need to be explicitly formed.
Therefore, we can apply the conjugate gradient (CG) method to solve \eqref{cgobj} by a sequence of matrix-vector products. 
This technique is called Hessian-free methods in optimization. Details of CG
methods in a Hessian-free Newton framework can be found
in, for example, Algorithm 2 of \cite{CJL07a}. 

\par Because the computational cost in \eqref{GV} is proportional to the number of instances, 
subsampled Hessian Newton methods have been proposed \citep{RHB11a, JM10a, CCW15a} to reduce the cost in solving the linear system \eqref{cgobj-LM}.    
They observe that the second term in \eqref{GapproxH} is the average training loss.
If the large number of data points are assumed to be from the same distribution, 
\eqref{GapproxH} can be reasonably approximated by selecting a subset $S \subset \{1,\ldots,l\}$
and having
\begin{equation*}
G^{S} = \frac{1}{C} \mathcal I + \frac{1}{|S|} \sum_{i \in S} (J^i)^T B^{i}J^i \approx G.
\end{equation*}
Then \eqref{GV} becomes
\begin{equation}
(G^{S} + \lambda \mathcal{I})\bv = (\frac{1}{C}+\lambda)\bv + \frac{1}{|S|} \sum_{i \in S} \left((J^i)^T \left(B^i (J^i \bv)\right)\right) \approx (G + \lambda \mathcal{I})\bv.
\label{Gv-reg}
\end{equation}

A summary of the Newton method is in Algorithm \ref{alg:newton-cnn}.
\renewcommand{\baselinestretch}{1.3}
\begin{algorithm}[t]
  \caption{A standard subsampled Hessian Newton method for CNN.}
  \begin{algorithmic}[1]
  \State Compute $f(\btheta^{1})$.
  \For{$k = 1,\ldots,$}
    \State Choose a set $S_k \subset \{1,\ldots,l\}$.
    \State Compute $\nabla f(\btheta^k)$ and $J^i, \forall i \in S_k$.
    \State Approximately solve the linear system in \eqref{cgobj-LM} by CG to obtain a direction $\bd^k$
    \State $\alpha = 1$.
    \While{true}
		\State Update $\btheta^{k+1} = \btheta^{k} + \alpha \bd^{k}$ and compute $f(\btheta^{k+1})$
		\If {\eqref{linesearch-condition} is satisfied}
			\State break
		\EndIf
        \State $\alpha \leftarrow \alpha/2$.
    \EndWhile
    \State Calculate $\lambda_{k+1}$ based on \eqref{LM-rules}.
  \EndFor
  \end{algorithmic}
  \label{alg:newton-cnn}
\end{algorithm}
\renewcommand{\baselinestretch}{2}

\subsection{Gradient Evaluation}
\label{subsec:Gradient}
In order to solve \eqref{cgobj}, $\nabla f(\btheta)$ is needed.
It can be obtained by \eqref{whole-gradient} if the Jacobian matrices $J^i,\
i=1,\ldots,l$ are available. From \eqref{GV}, it seems that $J^i, \forall i$ are
also needed for the matrix-vector product in CG. However, as mentioned in
Section \ref{subsec:proposed-alg}, in practice a sub-sampled Hessian method is
used, so from \eqref{Gv-reg} only a subset of $J^i, \forall i$ are needed. Therefore
we present a backward process to calculate the gradient without using Jacobian.

Consider two layers $\mfirst$ and $\msec$. The variables between them are $W^m$ and $\bb^m$, so we aim to calculate the following gradient components.
\begin{align}
\label{gradient-w}
\partialdiff{f}{W^m}
&= \frac{1}{C}W^{m} + \frac{1}{l} \sum^l_{i=1} \frac{\partial \xi_i}{\partial W^m},  \\
\label{gradient-b}
\partialdiff{f}{\bb^m}
&= \frac{1}{C}\bb^m + \frac{1}{l} \sum^l_{i=1} \partialdiff{\xi_i}{\bb^m}.
\end{align}
Because \eqref{gradient-w} is in a matrix form, following past developments such
as \cite{AV15a}, it is easier to transform them to a vector form for
the derivation. To begin, we list the following properties of the $\vec(\cdot)$
function, in which $\otimes$ is the Kronecker product.
\begin{align}
  \vec(AB)
  &= (\mathcal{I} \otimes A) \vec(B) \label{otimes-back},  \\
  &= (B^T \otimes \mathcal{I}) \vec(A) \label{otimes-front},  \\
  \vec(AB)^T
  &= \vec(B)^T (\mathcal{I} \otimes A^T) \label{otimes-transpose-front},  \\
  &= \vec(A)^T (B \otimes \mathcal{I}) \label{otimes-transpose-back}.
\end{align}

We further define
\begin{align}
  \mpartialdiff{\by}{(\bx)} &=
\begin{bmatrix}
\frac{\partial y_1}{\partial x_1} & \hdots & \frac{\partial y_1}{\partial x_{|\bx|}} \\
\vdots & \ddots & \vdots \\ 
\frac{\partial y_{|\by|}}{\partial x_1} & \hdots & \frac{\partial y_{|\by|}}{\partial x_{|\bx|}}  
\end{bmatrix}, \nonumber
\end{align}
where $\bx$ and $\by$ are column vectors.

For the convolutional layers, from
\eqref{conv-f-ztos} and Table \ref{operation-table}, we have
\begin{align}
  \vec(S^{\mfirst,i}) &= \vec(W^m \phi(\padding(Z^{\mfirst,i}))) + \vec(\bb^m
  \mathds{1}^T_{a^m_{\conv} b^m_{\conv}}) \nonumber\\
&= \left( \mathcal{I}_{a^{\mfirst}_{\conv} b^{\mfirst}_{\conv}} \otimes W^m \right)
\vec(\phi(\padding(Z^{\mfirst,i}))) + (\mathds{1}_{a^{\mfirst}_{\conv} b^{\mfirst}_{\conv}} \otimes
\mathcal{I}_{d^{\msec}}) \bb^m \label{conv-vec-z}\\
&= \left( \phi(\padding(Z^{\mfirst,i}))^T \otimes \mathcal{I}_{d^{\msec}} \right)
\vec(W^m) + (\mathds{1}_{a^{\mfirst}_{\conv} b^{\mfirst}_{\conv}} \otimes \mathcal{I}_{d^{\msec}}) \bb^m
  \label{conv-vec-w},
\end{align}
where \eqref{conv-vec-z} and \eqref{conv-vec-w} are from \eqref{otimes-back}
and \eqref{otimes-front}, respectively.

For the fully-connected layers, from \eqref{full-f-s}, we have
\begin{align}
\bs^{\mfirst,i} &= W^m \bz^{\mfirst,i} + \bb^m  \nonumber\\
          &= \left( \mathcal{I}_{1} \otimes W^m \right) \bz^{\mfirst,i}
  + (\mathds{1}_1 \otimes \mathcal{I}_{n_{\msec}})\bb^m  \label{full-vec-z}\\
  &= \left( (\bz^{\mfirst,i})^T \otimes \mathcal{I}_{n_{\msec}} \right) \vec(W^m)
  + (\mathds{1}_1 \otimes \mathcal{I}_{n_{\msec}})\bb^m,  \label{full-vec-w}
\end{align}
where \eqref{full-vec-z} and \eqref{full-vec-w} are from \eqref{otimes-back} and
\eqref{otimes-front}, respectively. 

An advantage of using \eqref{conv-vec-z} and \eqref{full-vec-z} is that they are
in the same form.
Further, if for fully-connected layers we define
\begin{equation*}
  \phi(\padding(\bz^{\mfirst,i})) = \mathcal{I}_{n_{\mfirst}} \bz^{\mfirst,i} ,\  L^c < m
  \leq \mout,
\end{equation*}
then \eqref{conv-vec-w} and \eqref{full-vec-w} are in the same form.
Thus we can derive the gradient together.
We begin with calculating the gradient for convolutional layers. From
\eqref{conv-vec-w}, we derive
\begin{align}
  \mpartialdiff{\xi_i}{\vec(W^m)}
  &= \mpartialdiff{\xi_i}{\vec(S^{\mfirst,i})}
  \mpartialdiff{\vec(S^{\mfirst,i})}{\vec(W^m)}  \nonumber\\
  &= \mpartialdiff{\xi_i}{\vec(S^{\mfirst,i})} \left(\phi(\padding(Z^{\mfirst,i}))^T\
\otimes\ \mathcal{I}_{d^{\msec}}\right)  \nonumber\\
  &= \vec\left(\partialdiff{\xi_i}{S^{\mfirst,i}} \phi(\padding(Z^{\mfirst,i}))^T \right)^T
\label{conv-dxidw}
\end{align}
and
\begin{align}
  \mpartialdiff{\xi_i}{(\bb^m)}
  &= \mpartialdiff{\xi_i}{\vec(S^{\mfirst,i})}
  \mpartialdiff{\vec(S^{\mfirst,i})}{(\bb^m)}  \nonumber\\
  &= \mpartialdiff{\xi_i}{\vec(S^{\mfirst,i})}\left(\mathds{1}_{a^{\mfirst}_{\conv}
b^{\mfirst}_{\conv}} \otimes \mathcal{I}_{d^{\msec}}\right)  \nonumber\\
  &= \vec\left(\partialdiff{\xi_i}{S^{\mfirst,i}}
\mathds{1}_{a^{\mfirst}_{\conv} b^{\mfirst}_{\conv}}\right)^T,
\label{conv-dxidb}
\end{align}
where \eqref{conv-dxidw} and \eqref{conv-dxidb} are from
\eqref{otimes-transpose-back}.
To calculate \eqref{conv-dxidw}, $\phi(\padding(Z^{\mfirst,i}))$ has been
available from the forward process of calculating the function value. In
\eqref{conv-dxidw} and \eqref{conv-dxidb}, $\partial \xi_i / \partial
S^{\mfirst,i}$ is also needed and can be obtained by a backward process. By
assuming that $\partial \xi_i / \partial Z^{\msec,i}$ is available, we show
details of calculating $\partial \xi_i / \partial S^{\mfirst,i}$ and $\partial
\xi_i / \partial Z^{m,i}$ for layer $m$. From \eqref{flow-conv} we have the following workflow.
\begin{equation}
Z^{\mfirst,i} \leftarrow \text{padding} \leftarrow
\text{convolution} \leftarrow \sigma(S^{\mfirst,i}) \leftarrow \text{pooling}
\leftarrow Z^{\msec,i}.
\label{gd:workflow}
\end{equation}

Because the RELU activation function is considered for the convolutional layers, we have
\begin{align}
\frac{\partial \xi_i}{\partial \vec(S^{\mfirst,i})^T} 
&= \frac{\partial \xi_i}{\partial \vec(Z^{\msec,i})^T} \mpartialdiff{\vec(Z^{\msec,i})}{\vec(\sigma(S^{\mfirst,i}))}
  \frac{\partial \vec(\sigma(S^{m,i}))}{\partial \vec(S^{m,i})^T} \label{gd:chain-xi-relu} \\
&= \left(\frac{\partial \xi_i}{\partial \vec(Z^{\msec,i})^T} \mpartialdiff{\vec(Z^{\msec,i})}{\vec(\sigma(S^{\mfirst,i}))} \right)\ \odot\ \vec(I[S^{\mfirst,i}])^T \label{gd:xi-relu} \\
&= \left(\frac{\partial \xi_i}{\partial \vec(Z^{\msec,i})^T}\ P^{m,i}_{\pool} \right)\ \odot\ \vec(I[S^{\mfirst,i}])^T, \label{gd:xi-z-s}
\end{align}
where \eqref{gd:xi-z-s} is from \eqref{pool-Z}, $\odot$ is Hadamard product (i.e., element-wise products), 
\begin{equation*}
I[S^{\mfirst,i}]_{(p,q)} = 
\begin{cases}
1 & \text{ if } s^{\mfirst,i}_{(p,q)} > 0,\\
0 & \text{ otherwise,}
\end{cases}
\end{equation*}
and because
\begin{equation*}
\frac{\partial \vec(\sigma(S^{m,i}))}{\partial \vec(S^{m,i})^T} 
\end{equation*}
is a diagonal matrix, \eqref{gd:xi-relu} can be derived from \eqref{gd:chain-xi-relu}.

Next, we must calculate $\partial \xi_i / \partial Z^{\mfirst,i}$ and pass it to the previous layer.
\begin{align}
 &\mpartialdiff{\xi_i}{\vec(Z^{\mfirst,i})} \nonumber \\
=& \mpartialdiff{\xi_i}{\vec(S^{\mfirst,i})}
   \mpartialdiff{\vec(S^{\mfirst,i})}{\vec(\phi(\padding(Z^{\mfirst,i})))}
   \frac{\partial \vec(\phi(\padding(Z^{m,i})))}{\partial \vec(\padding(Z^{m,i}))^T}
   \mpartialdiff{\vec(\padding(Z^{\mfirst,i}))}{\vec(Z^{\mfirst,i})} \nonumber\\
=& \mpartialdiff{\xi_i}{\vec(S^{\mfirst,i})}
  \left(\mathcal{I}_{a^{\mfirst}_{\conv}b^{\mfirst}_{\conv}} \otimes W^m
\right) P^{m}_{\phi} P^{m}_{\padding}\label{gd:dXidZ_kron} \\
=& \vec\left((W^m)^T \partialdiff{\xi_i}{S^{\mfirst,i}}\right)^T P^{m}_{\phi} P^{m}_{\padding},
\label{gd:dXidZ}
\end{align}
where \eqref{gd:dXidZ_kron} is from \eqref{fun-phi}, \eqref{padding} and \eqref{conv-vec-z}, and
\eqref{gd:dXidZ} is from \eqref{otimes-transpose-front}.

\par For fully-connected layers, by the same form in
\eqref{full-vec-z}, \eqref{full-vec-w}, \eqref{conv-vec-z} and \eqref{conv-vec-w},
we immediately get the following results from \eqref{conv-dxidw}, \eqref{conv-dxidb},
\eqref{gd:xi-z-s} and \eqref{gd:dXidZ}.
\begin{align}
  \mpartialdiff{\xi_i}{\vec(W^m)}
  &= \vec\left(\partialdiff{\xi_i}{\bs^{\mfirst,i}} (\bz^{\mfirst,i})^T\right)^T
\label{gd:full-w}\\
\mpartialdiff{\xi_i}{(\bb^m)}
&= \mpartialdiff{\xi_i}{(\bs^{\mfirst,i})}  \label{gd:full-b}\\
\mpartialdiff{\xi_i}{(\bz^{\mfirst,i})}
&= \left( (W^m)^T \partialdiff{\xi_i}{(\bs^{\mfirst,i})} \right) ^T
\mathcal{I}_{n_{\mfirst}}
\nonumber\\
&= \mpartialdiff{\xi_i}{(\bs^{\mfirst,i})}
W^m,  \label{gd:full-z}
\end{align}
where
\begin{equation}
\label{gd:full-s}
\mpartialdiff{\xi_i}{(\bs^{\mfirst,i})} = \mpartialdiff{\xi_i}{(\bz^{\msec,i})} \odot
I[\bs^{\mfirst,i}]^T. 
\end{equation}
Finally, we check the initial values of the backward process. From the square
loss in \eqref{square-loss} and the linear activation function in \eqref{linear}, we have
\begin{align*}
  \partialdiff{\xi_i}{\bz^{\mout,i}} &= 2 ( \bz^{\mout,i} - \by^i ),  \\
  \partialdiff{\xi_i}{\bs^{L,i}} &= \partialdiff{\xi_i}{\bz^{\mout,i}}.
\end{align*}

\subsubsection{Some Notes on Practical Implementations}
\label{subsubsec:gd-padding-pooling}

In practice, because we only store
\begin{equation*}
  Z^{\msec,i} = \mat\left(P^{m,i}_{\pool} \vec(\sigma(S^{\mfirst,i}))\right)
\end{equation*}
rather than $\sigma(S^{\mfirst,i})$, instead of using \eqref{gd:xi-z-s}, we conduct the following calculation.
\begin{equation}
	\mpartialdiff{\xi_i}{\vec(S^{\mfirst,i})}
= \left( \mpartialdiff{\xi_i}{\vec(Z^{\msec,i})} \odot \vec(I[Z^{\msec,i}])^T \right) P^{m,i}_{\pool}.
\label{padding-without-pooling:vTP}
\end{equation}
The reason is that, for \eqref{gd:xi-z-s},
\begin{equation}
	\mpartialdiff{\xi_i}{\vec(Z^{\msec,i})}  \times P^{m,i}_{\pool}
\label{dxidz_pool}
\end{equation}
generates a large zero vector and puts values of $\partial
\xi_i/\partial \vec(Z^{\msec,i})^T$ into positions selected earlier in the max pooling operation.
Then, element-wise multiplications of \eqref{dxidz_pool} and $I[S^{\mfirst,i}]^T$ follow.
Because positions not selected in the earlier max pooling procedure in \eqref{dxidz_pool} are zeros 
and they are still zeros after the Hadamard product between \eqref{dxidz_pool} and
$I[S^{\mfirst,i}]^T$, \eqref{gd:xi-z-s} and \eqref{padding-without-pooling:vTP} give the same results.

\subsection{Jacobian Evaluation}
\label{subsec:Jacobian}
For the matrix-vector product \eqref{GapproxH}, the Jacobian matrix is needed.
We note that it can be partitioned into $L$ blocks.
\begin{equation}
\label{J-split}
  J^i = \begin{bmatrix}
    J^{1,i} & J^{2,i} & \hdots & J^{L,i}
  \end{bmatrix},\ m=1,\ldots,L,\ i=1,\ldots,l,
\end{equation}
where
\begin{equation*}
J^{m,i} = \left[\frac{\partial \bz^{\mout,i}}{\partial \vec(W^m)^T}\arrayspace
\frac{\partial \bz^{\mout,i}}{\partial (\bb^m)^T}\right].
\end{equation*}

The calculation is very similar to that for the gradient. For the convolutional layers, from \eqref{conv-dxidw} and \eqref{conv-dxidb}, we have
\begin{align}
\left[\frac{\partial \bz^{\mout,i}}{\partial \vec(W^m)^T}\arrayspace
\frac{\partial \bz^{\mout,i}}{\partial (\bb^m)^T} \right]
&=
\begin{bmatrix}
\frac{\partial z^{\mout,i}_{1}}{\partial \vec(W^m)^T} \arrayspace \frac{\partial
z^{\mout,i}_{1}}{\partial (\bb^m)^T} \\
\vdots \\
\frac{\partial z^{\mout,i}_{n_\mout}}{\partial \vec(W^m)^T} \arrayspace
\frac{\partial z^{\mout,i}_{n_\mout}}{\partial (\bb^m)^T}
\end{bmatrix} \nonumber\\
&=
\begin{bmatrix}
\vec(\frac{\partial z^{\mout,i}_{1}}{\partial S^{\mfirst,i}} \phi(\padding(Z^{\mfirst,i}))^T)^T
\arrayspace \vec(\frac{\partial z^{\mout,i}_{1}}{\partial S^{\mfirst,i}}
\mathds{1}_{\convaout \convbout})^T \\
\vdots \\
\vec(\frac{\partial z^{\mout,i}_{n_\mout}}{\partial S^{\mfirst,i}}
\phi(\padding(Z^{\mfirst,i}))^T)^T \arrayspace \vec(\frac{\partial
z^{\mout,i}_{n_\mout}}{\partial S^{\mfirst,i}} \mathds{1}_{\convaout \convbout})^T \\
\end{bmatrix} \nonumber\\
&=
\begin{bmatrix}
\vec\left(\frac{\partial z^{\mout,i}_1}{\partial S^{\mfirst,i}}
\left[\phi(\padding(Z^{\mfirst,i}))^T\ \mathds{1}_{\convaout \convbout}\right]\right)^T \\
\vdots\\
\vec\left(\frac{\partial z^{\mout,i}_{n_\mout}}{\partial S^{\mfirst,i}}
\left[\phi(\padding(Z^{\mfirst,i}))^T\ \mathds{1}_{\convaout \convbout}\right]\right)^T
\end{bmatrix}.
\label{jaco:conv-w} 
\end{align}

We use a backward process to calculate $\partial \bz^{L+1,i}/\partial S^{m,i},\ \forall i$. 
Assume that $\partial \bz^{\mout,i}/\partial Z^{\msec,i}$ are available. 
From \eqref{padding-without-pooling:vTP}, we have
\begin{equation*}
\frac{\partial z^{\mout,i}_j}{\partial \vec(S^{\mfirst,i})^T} = \left( \frac{\partial
z^{\mout,i}_j}{\partial \vec(Z^{\msec,i})^T} \odot \vec(I[Z^{\msec,i}])^T \right) P^{m,i}_{\pool},\
j=1,\ldots,n_\mout.
\end{equation*}
These vectors can be written together as
\begin{equation}
\mpartialdiff{\bz^{\mout,i}}{\vec(S^{\mfirst,i})}
= \left(\mpartialdiff{\bz^{\mout,i}}{\vec(Z^{\msec,i})} \odot
\left(\mathds{1}_{n_\mout} \vec(I[Z^{\msec,i}])^T \right)\right) P^{m,i}_{\pool}.
\label{jaco:xi-z-s}
\end{equation}

We then generate $\partial \bz^{\mout,i} / \partial \vec(Z^{\mfirst,i})^T$ and pass it to the previous layer.
From \eqref{gd:dXidZ}, we derive
\begin{align}
\frac{\partial \bz^{\mout,i}}{\partial \vec(Z^{\mfirst,i})^T} &=
\begin{bmatrix}
\frac{\partial z^{\mout,i}_{1}}{\partial \vec(Z^{\mfirst,i})^T} \\
\vdots \\
\frac{\partial z^{\mout,i}_{n_\mout}}{\partial \vec(Z^{\mfirst,i})^T}
\end{bmatrix} \nonumber\\
&=
\begin{bmatrix}
\vec\left((W^m)^T \frac{\partial z^{\mout,i}_{1}}{\partial S^{\mfirst,i}}\right)^T P^{m}_{\phi} P^m_{\padding}\\
\vdots\\
\vec\left((W^m)^T \frac{\partial z^{\mout,i}_{n_\mout}}{\partial S^{\mfirst,i}}\right)^T P^{m}_{\phi} P^m_{\padding}
\end{bmatrix}.
\label{jaco:zL-z}
\end{align}

For the fully-connected layers, we follow the same derivation of gradient to
have
\begin{align}
\mpartialdiff{\bz^{\mout,i}}{\vec(W^m)}
&= \left[ \vec( \partialdiff{z^{\mout,i}_1}{\bs^{\mfirst,i}} (\bz^{\mfirst,i})^T
) \ \ \hdots\ \ \vec( \partialdiff{z^{\mout,i}_{n_\mout}}{\bs^{\mfirst,i}}
(\bz^{\mfirst,i})^T )
\right]^T,  \label{jaco:xi-w-full}\\
\frac{\partial \bz^{\mout,i}}{\partial (\bb^m)^T}
&= \frac{\partial \bz^{\mout,i}}{\partial (\bs^{\mfirst,i})^T}, \label{jaco:xi-b-full}\\
\frac{\partial \bz^{\mout,i}}{\partial (\bz^{\mfirst,i})^T}
&= \frac{\partial \bz^{\mout,i}}{\partial (\bs^{\mfirst,i})^T}  W^m,  \label{jaco:xi-z-full}\\
\frac{\partial \bz^{\mout,i}}{\partial (\bs^{\mfirst-1,i})^T}
&= \frac{\partial \bz^{\mout,i}}{\partial (\bz^{\mfirst,i})^T} \odot
\left(\mathds{1}_{n_\mout} I[\bz^{\mfirst,i}]^T\right).
\label{jaco:xi-z-s-full}
\end{align}
For layer $\mout$, because
of using \eqref{square-loss} and the linear activation function, we have
\begin{equation*}
  \frac{\partial \bz^{\mout,i}}{\partial (\bs^{L,i})^T} =
  \mathcal{I}_{n_\mout}.
\end{equation*}

\subsection{Gauss-Newton Matrix-Vector Products}
\label{subsec:GN-matrix-vector}
Conjugate gradient (CG) methods are used to solve \eqref{cgobj}.
The main operation at each CG iteration is the Gauss-Newton matrix-vector
product in \eqref{GV} or \eqref{Gv-reg} with the subsampled setting.

From \eqref{J-split}, we rearrange \eqref{GapproxH} to 
\begin{align}
G = \frac{1}{C}\mathcal{I} + \frac{1}{l}\sum_{i=1}^l
\begin{bmatrix}
(J^{1,i})^T \\ \vdots \\ (J^{L,i})^T
\end{bmatrix}
B^i
\begin{bmatrix}
J^{1,i} & \hdots & J^{L,i}
\end{bmatrix}
\label{Gv:G-expand}
\end{align}
and the Gauss-Newton matrix vector product becomes
\begin{align}
G\bv
&= \frac{1}{C} \bv + \frac{1}{l}
\sum_{i=1}^l 
\begin{bmatrix}
(J^{1,i})^T \\ \vdots \\ (J^{L,i})^T
\end{bmatrix}
B^i
\begin{bmatrix}
J^{1,i} & \hdots & J^{L,i}
\end{bmatrix}
\begin{bmatrix}
\bv^1 \\ \vdots \\ \bv^{L}
\end{bmatrix} \nonumber \\
&=
\frac{1}{C} \bv + \frac{1}{l}
\sum_{i=1}^l 
\begin{bmatrix}
(J^{1,i})^T \\ \vdots \\ (J^{L,i})^T
\end{bmatrix}
\left(B^i \sum_{m=1}^L J^{m,i} \bv^m \right),
\label{Gv:block-diag}
\end{align}
where
\begin{equation*}
  \bv = \begin{bmatrix}
    \bv^1 \\ \vdots \\ \bv^L
  \end{bmatrix},
\end{equation*}
and each $\bv^m, m = 1, \hdots, L$ has the same length as the number of
variables (including bias) at the $m$th layer.

\par For the convolutional layers, from \eqref{jaco:conv-w} and
\eqref{Gv:block-diag}, we have
\begin{equation}
J^{m,i}\bv^m =
\begin{bmatrix}
 \vec\left(\frac{\partial z^{\mout,i}_1}{\partial S^{\mfirst,i}}
   \left[\phi(\padding(Z^{\mfirst,i}))^T\
 \mathds{1}_{\convaout \convbout}\right]\right)^T \bv^m\\
\vdots\\
 \vec\left(\frac{\partial z^{\mout,i}_{n_\mout}}{\partial S^{\mfirst,i}}
 \left[\phi(\padding(Z^{\mfirst,i}))^T\ \mathds{1}_{\convaout \convbout}\right]\right)^T \bv^m
\end{bmatrix}.
\label{Jv-ori}
\end{equation}
To simplify \eqref{Jv-ori}, we use the following property
\begin{equation*}
\vec(AB)^T \vec(C) = \vec(A)^T \vec(CB^T)
\end{equation*}
to have that for example, the first element in \eqref{Jv-ori} is
\begin{align*}
  &\ \vec\left(\frac{\partial z^{\mout,i}_1}{\partial S^{\mfirst,i}}
\left[\phi(\padding(Z^{\mfirst,i}))^T\ \mathds{1}_{\convaout \convbout}\right]\right)^T \bv^m \\
 =&\ \frac{\partial z^{\mout,i}_{1}}{\partial \vec(S^{\mfirst,i})^T} \vec\left(
 \mat(\bv^m)_{d^\msec \times (h^mh^md^\mfirst + 1)} \begin{bmatrix}
 \phi(\padding(Z^{\mfirst,i})) \\  \mathds{1}^T_{\convaout \convbout} \end{bmatrix}\right).
\end{align*}
Therefore,
\begin{equation}
J^{m,i}\bv^m =
\frac{\partial \bz^{\mout,i}}{\partial \vec(S^{\mfirst,i})^T} \vec\left(
\mat(\bv^m)_{d^\msec \times (h^mh^md^\mfirst + 1)} \begin{bmatrix}
\phi(\padding(Z^{\mfirst,i})) \\  \mathds{1}^T_{\convaout \convbout} \end{bmatrix}\right).
\label{Jv}
\end{equation}

After deriving \eqref{Jv}, from \eqref{Gv:block-diag}, we sum results of all layers
\begin{equation*}
\sum^L_{m=1} J^{m,i}\bv^m
\end{equation*}
and then calculate
\begin{equation}
  \bq^i = B^i ( \sum^L_{m=1} J^{m,i}\bv^m).
\label{BJv-inst}
\end{equation}
From \eqref{square-loss} and \eqref{Bts},  
\begin{equation}
B^i_{ts} = \frac{\partial^2 \xi^i}{\partial z^{\mout,i}_t \partial z^{\mout,i}_s}
= \frac{\partial^2 (\sum_{j=1}^{n_\mout} (z^{\mout,i}_j - y^{i}_j)^2 )}{\partial
z^{\mout,i}_t \partial z^{\mout,i}_s}
=
\begin{cases}
2 & \text{if } t = s,\\
0 & \text{otherwise,}
\end{cases}
\label{B-def}
\end{equation}
and we derive $\bq^i$ by multiplying every element of $\sum^L_{m=1} J^{m,i}\bv^m$ by two.

\par After deriving \eqref{BJv-inst}, from \eqref{jaco:conv-w} and \eqref{Gv:block-diag}, we calculate
\begin{align}
&\ (J^{m,i})^T \bq^{i} \nonumber \\
=&\ \begin{bmatrix}
\vec\left(\frac{\partial z^{\mout,i}_1}{\partial S^{\mfirst,i}}
\left[\phi(\padding(Z^{\mfirst,i}))^T\ \mathds{1}_{\convaout \convbout}\right]\right) &
\cdots &
\vec\left(\frac{\partial z^{\mout,i}_{n_\mout}}{\partial S^{\mfirst,i}}
\left[\phi(\padding(Z^{\mfirst,i}))^T\ \mathds{1}_{\convaout \convbout}\right]\right)
\end{bmatrix} \bq^{i} \nonumber \\
=&\ 
 \sum_{j=1}^{n_\mout} q^{i}_j \vec\left( \frac{\partial
 z^{\mout,i}_{j}}{\partial S^{\mfirst,i}} \left[\phi(\padding(Z^{\mfirst,i}))^T\
 \mathds{1}_{\convaout \convbout}\right]\right) \nonumber \\
=&\ 
\vec\left( \sum_{j=1}^{n_\mout} q^{i}_j \left(\frac{\partial
z^{\mout,i}_{j}}{\partial S^{\mfirst,i}}\left[\phi(\padding(Z^{\mfirst,i}))^T\
\mathds{1}_{\convaout \convbout}\right]\right) \right) \nonumber \\
=&\ 
\vec\left( \left(\sum_{j=1}^{n_\mout} q^{i}_j \frac{\partial
z^{\mout,i}_{j}}{\partial S^{\mfirst,i}}\right) \left[\phi(\padding(Z^{\mfirst,i}))^T\
\mathds{1}_{\convaout \convbout}\right]\right) \nonumber \\
=&\ 
\vec\left( \mat\left(\left(\frac{\partial \bz^{\mout,i}}{\partial
\vec(S^{\mfirst,i})^T}\right)^T\bq^{i}\right)_{\convdout \times \convaout \convbout}
\left[\phi(\padding(Z^{\mfirst,i}))^T\ \mathds{1}_{\convaout \convbout}\right]\right).
\label{JtV}
\end{align}

Similar to the results of the convolutional layers, for the fully-connected layers we have
\begin{align}
J^{m,i}\bv^m 
&= \frac{\partial \bz^{\mout,i}}{\partial (\bs^{\mfirst,i})^T}
\mat(\bv^m)_{n_\msec \times (n_{\mfirst}+1)} \begin{bmatrix} \bz^{\mfirst,i} \\ \mathds{1}_{1} \end{bmatrix}.
\label{full-Jv}
\end{align}
\begin{align}
(J^{m,i})^T \bq^{i}
&= \vec\left(\left(\frac{\partial \bz^{\mout,i}}{\partial
(\bs^{\mfirst,i})^T}\right)^T \bq^{i} \left[ (\bz^{\mfirst,i})^T\ \mathds{1}_{1}\right]\right).
\label{full-JtV}
\end{align}

\section{Implementation Details}
\label{sec:impl}
We show that with a careful design, a Newton method for CNN can be implemented by a simple and short program.
A {\sl MATLAB} implementation is given as an illustration though modifications for other languages such as {\sl Python}
should be straightforward.
\par For the discussion in Section \ref{sec:HessianFree}, we check each
individual data. However, for practical implementations, all instances must be
considered together for memory and computational efficiency. In our
implementation, we store $Z^{\mfirst,i},\ \forall i=1, \hdots, l$ as the following
matrix.
\begin{equation}
\label{impl:Z}
  \begin{bmatrix}
    Z^{\mfirst, 1} & Z^{\mfirst, 2} & \hdots & Z^{\mfirst, l}
  \end{bmatrix} \in
  R^{\paddin \times \padain \padbin l}.
\end{equation}
Similarly, we store $\partial \xi_{i} / \partial \vec(S^{\mfirst,i})^T,\ \forall i$ as
\begin{equation}
  \label{impl:dXidS}
  \begin{bmatrix}
    \partialdiff{\xi_1}{S^{\mfirst,1}} & \hdots &
    \partialdiff{\xi_l}{S^{\mfirst,l}}
  \end{bmatrix} \in
  R^{\pooldin \times \poolain \poolbin l}.
\end{equation}
For $\partial \bz^{\mout,i} / \partial \vec(S^{\mfirst,i})^T,\ \forall i$, we
consider
\begin{equation}
  \begin{bmatrix}
    \partialdiff{z^{\mout,1}_{1}}{S^{\mfirst, 1}} & \hdots &
    \partialdiff{z^{\mout,1}_{n_\mout}}{S^{\mfirst, 1}} & \hdots &
    \partialdiff{z^{\mout,l}_{n_\mout}}{S^{\mfirst, l}}
  \end{bmatrix} \in
  R^{\pooldin \times \poolain \poolbin n_\mout l}
\label{impl:dzdS}
\end{equation}
and will explain our decision. Note that \eqref{impl:Z}-\eqref{impl:dzdS} are
only the main setting to store these matrices because for some operations they
may need to be re-shaped.

For an easy description in some places we follow Section \ref{subsec:conv-layer} to let
\begin{equation*}
Z^{\inpu,i} \text{ and } Z^{\oupu,i}
\end{equation*}
be the input and output images of a layer, respectively.

\subsection{Generation of $\phi(\padding(Z^{m,i}))$}
\label{subsec:impl-phi}
{\sl MATLAB} has a built-in function \im2col that can generate
$\phi(\padding(Z^{m,i}))$ for $s = 1$ and $s = h$.
For general $s$, we notice that $\phi(\padding(Z^{m,i}))$ is a sub-matrix of the output matrix of using {\sl MATLAB}'s
\im2col under $s = 1$. In supplementary materials we provide an efficient implementation to extract the sub-matrix.
However, this approach is not ideal because first in other languages a subroutine
like {\sl MATLAB}'s \im2col may not be available, and second, generating a larger matrix under $s = 1$
causes extra time and memory. 

Therefore, here we show an efficient implementation without relying on a
subroutine like {\sl MATLAB}'s \im2col. For an easy description we follow
Section \ref{subsec:conv-layer} to consider
\begin{equation*}
\padding(Z^{m,i}) = Z^{\inpu, i} \rightarrow Z^{\oupu, i} = \phi(Z^{\inpu, i}).
\end{equation*}
Consider the following linear indices\footnote{Linear indices refer to the
  sequence of how elements in a matrix are stored. Here we consider a
column-oriented setting.} (i.e., counting elements in a column-oriented way) of
$\genphiZin$:
\begin{equation}
  \label{zm-linearidx}
\begin{bmatrix}
  1  &  \din + 1  & \hdots  &  (\bin \ain - 1) \din + 1  \\
  2  &  \din + 2  & \hdots  &  (\bin \ain - 1) \din + 2  \\
  \vdots  &  \vdots  & \ddots  &  \vdots \\
  \din &  2\din & \hdots  &  (\bin \ain) \din
\end{bmatrix} \in R^{\din \times \ain \bin}.
\end{equation}
Because every element in
\begin{equation*}
  \phi(\genphiZin) \in R^{\genphih \genphih \din \times \aout \bout},
\end{equation*}
is extracted from $\genphiZin$, the task is to find the mapping between each
element in $\phi(\genphiZin)$ and a linear index of $\genphiZin$.
Consider the following example.
\begin{equation*}
\ain = 3,\ \bin = 2,\ \din = 1,\ s = 1,\ h = 2.
\end{equation*}
Because $\din=1$, we omit the channel subscript. In addition, we omit the
instance index $i$, so the image is
\begin{equation*}
\begin{bmatrix}
z_{11} & z_{12} \\
z_{21} & z_{22} \\
z_{31} & z_{32}
\end{bmatrix}.
\end{equation*}
By our representation in \eqref{zm-1},
\begin{equation*}
  Z^{\inpu} = \begin{bmatrix}
    z_{11} & z_{21} & z_{31} & z_{12} & z_{22} & z_{32}
  \end{bmatrix}
\end{equation*}
and the linear indices from \eqref{zm-linearidx} are
\begin{equation*}
  \begin{bmatrix}
    1 & 2 & 3 & 4 & 5 & 6
  \end{bmatrix}.
\end{equation*}
From \eqref{big-zm-1},
\begin{equation*}
\phi(Z^{\inpu}) =
\begin{bmatrix}
z_{11} & z_{21} \\
z_{21} & z_{31} \\
z_{12} & z_{22} \\
z_{22} & z_{32}
\end{bmatrix}.
\end{equation*}
Thus we store the following vector to indicate the mapping between linear
indices of $Z^{\inpu}$ and elements in $\phi(Z^{\inpu})$.
\begin{equation}
  \begin{bmatrix}
    1 & 2 & 4 & 5 & 2 & 3 & 5 & 6
  \end{bmatrix}^T.
\label{linearidx-eg}
\end{equation}
It also corresponds to column indices of non-zero elements in $P^{m}_{\phi}$.

To have a general setting we begin with checking how linear indices of $\genphiZin$ can be mapped to the
first column of $\phi(\genphiZin)$. For simplicity, we consider only channel $j$.
From $\eqref{big-zm-1}$ and \eqref{zm-linearidx}, we have
\begin{align}
\label{bigZ-linearidx}
\begin{bmatrix}[c|c]
j  &  z^{\inpu}_{1,1,j}  \\
\din + j  &  z^{\inpu}_{2,1,j}  \\
\vdots  &  \vdots \\
(\genphih-1) \din + j  &  z^{\inpu}_{\genphih,1,j}  \\
\ain \din + j  &  z^{\inpu}_{1,2,j}  \\
\vdots  &  \vdots  \\
((\genphih - 1) + \ain) \din + j  &  z^{\inpu}_{\genphih,2,j} \\
\vdots  &  \vdots  \\
((\genphih-1) + (\genphih - 1) \ain) \din + j  & z^{\inpu}_{\genphih,\genphih,j}  \\
\end{bmatrix},
\end{align}
where the left column gives the linear indices in $\genphiZin$,
while the right column shows the corresponding values.
We rewrite linear indices in \eqref{bigZ-linearidx} as
\begin{equation}
    \begin{bmatrix}
      0 + 0 \ain  \\
      \vdots  \\
      (\genphih-1) + 0 \ain  \\
      0 + 1 \ain  \\
      \vdots  \\
      (\genphih-1) + 1 \ain  \\
      \vdots  \\
      0 + (\genphih-1) \ain  \\
      \vdots  \\
      (\genphih-1) + (\genphih-1) \ain  \\
    \end{bmatrix} \din + j.
\label{linearidx-rewrite}
\end{equation}
Clearly, every linear index in \eqref{linearidx-rewrite} can be represented as
\begin{equation}
  \label{bigZ-linearidx-gen}
  (p + q \ain) \din + j,
\end{equation}
where
\begin{equation*}
p,\ q \in \{0,\ldots,\genphih -1\}
\end{equation*}
correspond to the pixel position in the convolutional filter.\footnote{
More precisely, $p+1$ and $q+1$ are the pixel position.
}

Next we consider other columns in $\phi(\genphiZin)$ by still fixing the channel
to be $j$. From \eqref{big-zm-1}, similar to the right column in
\eqref{bigZ-linearidx}, each column contains the following elements from the
$j$th channel of $\genphiZin$.
\begin{align}
  \label{bigZ-subscript}
  \genphizin_{1 + p + a \genphis, 1 + q + b \genphis, j}\ ,\ \ &a = 0, 1, \hdots,
  \aout-1,
  \nonumber  \\
  &b = 0, 1, \hdots, \bout-1,
\end{align}
where $(1 + a \genphis,\ 1 + b \genphis)$ denotes the top-left position of a sub-image in
the channel $j$ of $\genphiZin$. From \eqref{zm-linearidx}, the linear index of each
element in \eqref{bigZ-subscript} is
\begin{align}
  &((1+p + a\genphis - 1) + (1+q+b\genphis - 1) \ain)\din + j \nonumber\\
  =\ &(a+b\ain)\genphis\din + \underbrace{(p + q\ain)\din + j}_{\text{see
\eqref{bigZ-linearidx-gen}}}.
\label{all-mapped-idx}
\end{align}

Now we have known for each element of $\phi(\genphiZin)$ what the corresponding
linear index in $\genphiZin$ is. Next we discuss the implementation details,
where the code is shown in Listing \ref{list:phi}.
First, we compute elements in \eqref{linearidx-rewrite} with $j=1$ by applying
{\sl MATLAB}'s `{\tt +}' operator, which has the implicit expansion behavior, to
compute the outer sum of the following two arrays.
\begin{equation*}
  \begin{bmatrix}
    1  \\
    \din + 1  \\
    \vdots  \\
    (\genphih - 1) \din + 1  \\
  \end{bmatrix}
  \quad \text{and} \quad
  \begin{bmatrix}
    0  &  \ain \din  & \hdots  & (\genphih - 1) \ain \din
  \end{bmatrix}.
\end{equation*}
The result is the following matrix
\begin{equation}
  \label{first-channel-idx}
  \begin{bmatrix}
    1  &  \ain \din + 1  &  \hdots  &  (\genphih-1) \ain
    \din + 1\\
    \din + 1  &  (1 + \ain) \din + 1  &  \hdots  &  (1 +
    (\genphih-1) \ain) \din + 1 \\
    \vdots  & \vdots  &  \hdots & \vdots \\
    (\genphih - 1) \din + 1  &  ((\genphih-1) + \ain) \din + 1  & \hdots  &
    ((\genphih-1) + (\genphih-1) \ain) \din + 1\\
  \end{bmatrix},
\end{equation}
whose columns, if concatenated, lead to values in \eqref{linearidx-rewrite} with
$j=1$; see line \ref{list:phi|fch} of the code.
To get \eqref{bigZ-linearidx-gen} for all channels $j = 1, \hdots, \din$,
we compute the outer sum of the vector form of \eqref{first-channel-idx}
and
\begin{equation*}
  \begin{bmatrix}
    0 & 1 & \hdots & \din-1
  \end{bmatrix},
\end{equation*}
and then vectorize the resulting matrix; see line \ref{list:phi|fcol}.

To obtain other columns in $\phi(\genphiZin)$, we first calculate 
$\aout$ and $\bout$ by \eqref{abs} in lines \ref{list:phi|outa}-\ref{list:phi|outb}.
In the linear indices in \eqref{all-mapped-idx}, the second term
corresponds to indices of the first column, while the first term is the
following column offset
\begin{align*}
  (a + b \ain) \genphis \din,\ \forall &a = 0, 1, \hdots, \aout-1,\\
                                           &b = 0, 1, \hdots, \bout-1.
\end{align*}
This is the outer sum of the following two arrays.
\begin{equation*}
  \begin{bmatrix}
    0 \\ \vdots \\ \aout - 1
  \end{bmatrix} \times \genphis \din
  \quad\text{and}\quad \begin{bmatrix}
    0 & \hdots & \bout - 1
  \end{bmatrix} \times \ain \genphis \din;
\end{equation*}
see line \ref{list:phi|co} in the code. Finally, we compute the outer sum of the column
offset and the linear indices in the first column of $\phi(Z^{\inpu,i})$; see
line \ref{list:phi|idx}. In the end what we keep is the following vector
\begin{equation}
  \begin{bmatrix}
    \text{Column index of non-zero} \\
    \text{in each row of } P^{m}_{\phi}
  \end{bmatrix}_{h h \din \aout \bout}.
  \label{idxphi}
\end{equation}
Note that each row in the $0/1$ matrix $P^{m}_{\phi}$ contains exactly only one
non-zero element. We also see that \eqref{linearidx-eg} is an example of \eqref{idxphi}.

\par The obtained linear indices are independent of the values of $Z^{\inpu,i}$. Thus the above procedure
only needs to be run once in the beginning. For any $Z^{\inpu,i}$, we apply the
indices in \eqref{idxphi}
to extract $\phi(Z^{\inpu,i})$;
see line \ref{list:phi|extract-phiZ-st}-\ref{list:phi|extract-phiZ-ed}
in Listing \ref{list:phi}.
\par For the pooling operation $\phi(\genphiZin)$ is needed in \eqref{pool:phi}. The same implementation can be used.

\renewcommand{\baselinestretch}{1.3}
\begin{lstlisting}[mathescape=true,caption = {{\sl MATLAB} implementation for
$\phi(\genphiZin)$},label={list:phi},float=t]
function idx = find_index_phiZ(a,b,d,h,s)

first_channel_idx = ([0:h-1]*d+1)' + [0:h-1]*a*d; %( \label{list:phi|fch} %)
first_col_idx = first_channel_idx(:) + [0:d-1]; %( \label{list:phi|fcol} %)
out_a = floor((a - h)/s) + 1;  %( \label{list:phi|outa} %)
out_b = floor((b - h)/s) + 1;  %( \label{list:phi|outb} %)
column_offset = ([0:out_a-1]' + [0:out_b-1]*a)*s*d; %( \label{list:phi|co} %)
idx = column_offset(:)' + first_col_idx(:); %( \label{list:phi|idx} %)
idx = idx(:);


function phiZ = padding_and_phiZ(model, net, m)

a = model.ht_input(m);
b = model.wd_input(m);
num_data = size(net.Z{m},2)/a/b;

phiZ = padding(model, net, m);

% Calculate phiZ
phiZ = reshape(phiZ, [], num_data);  %( \label{list:phi|extract-phiZ-st} %)
phiZ = phiZ(net.idx_phiZ{m}, :);     %( \label{list:phi|extract-phiZ-ed} %)

h = model.wd_filter(m);
d = model.ch_input(m);
phiZ = reshape(phiZ, h*h*d, []);
\end{lstlisting}
\renewcommand{\baselinestretch}{2}

\subsection{Construction of $P^{m,i}_{\pool}$}
\label{subsec:Pm-1-pool}
Following \eqref{simple-maxpooling}, we use $Z^{\inpu,i}$ and $Z^{\oupu,i}$ to
represent the input
\begin{align*}
  \sigma(S^{\mfirst,i}) \in R^{\pooldin \times \poolain \poolbin}
\end{align*}
and the output
\begin{align*}
  Z^{\msec,i} \in R^{\pooldout \times \poolaout \poolbout}
\end{align*}
of the pooling operation, respectively. We need to store
$P^{m,i}_{\pool}$ because, besides function evaluations, it is used in gradient
and Jacobian evaluations; see \eqref{gd:xi-z-s} and \eqref{jaco:xi-z-s}.\footnote{Note
  that we do not really generate a sparse matrix $P^{m,i}_{\pool}$ in
\eqref{simple-maxpooling}. We only store column indices of non-zeros in
$P^{m,i}_{\pool}$.}
From \eqref{P_pool}, we need both $P^{m,i}_{\phi}$ and $W^{m,i}$. Because
$P^{m,i}_{\phi}$ is for partitioning each image to non-overlapping sub-regions
in \eqref{pool:phi} and \eqref{pool:output-a-b},
it is iteration independent. We obtain it in the beginning of the training
procedure by the method in Section \ref{subsec:impl-phi}.
\par For $W^{m,i}$, it is iteration dependent because the maximal value of each
sub-image is not a constant. Therefore, we construct
\begin{equation*}
  P^{m,i}_{\pool} = W^{m,i} P^{m,i}_{\phi} \in R^{\pooldout \poolaout \poolbout
  \times \pooldin \poolain \poolbin}
\end{equation*}
at the beginning of each Newton iteration, where a {\sl MATLAB} implementation is given
in Listing \ref{list:p-pool}.

To begin, we get
\begin{equation}
  Z^{\inpu,i},\ i = 1,\ldots,l,
\label{impl:z-dim}
\end{equation}
which are stored as a matrix in \eqref{impl:Z}.
Because \eqref{pool:output-a-b} may not hold with $a^{\oupu}$ and $b^{\oupu}$
being integers, we consider a setting the same as \eqref{abs}.
In line \ref{list:p-pool|P}, we extract the linear indices of $Z^{\inpu,i}$ to
appear in $\vec(\phi(Z^{\inpu,i}))$, which as we mentioned has been generated in
the beginning of the training procedure. The resulting vector {\tt P} contains
\begin{equation*}
h h \pooldin \poolaout \poolbout
\end{equation*}
elements and each element is in the range of
\begin{equation*}
1,\ \ldots,\ \pooldin \poolain \poolbin.
\end{equation*}
In line \ref{list:p-pool|gen-vec-phi-st}-\ref{list:p-pool|gen-vec-phi-ed}, we
use {\tt P} to generate
\begin{equation}
\label{pool:phiz}
\left[\vec(\phi(Z^{\inpu,1}))\ \cdots\ \vec(\phi(Z^{\inpu,l}))\right]\ \in
R^{h h \pooldin \poolaout \poolbout \times l}.
\end{equation}
Next we rewrite the above matrix so that each column contains a sub-region:
\begin{equation}
\label{pool:matphiz}
\begin{bmatrix}
z^{\mfirst,1}_{1,1,1} & z^{\mfirst,1}_{1,1,2} & \ldots &
  z^{\mfirst,l}_{1+(\poolaout - 1)\times s,1+(\poolbout - 1)\times
  s,\pooldin} \\
\vdots & \vdots & \ddots & \vdots \\
z^{\mfirst,1}_{h,h,1} & z^{\mfirst,1}_{h,h,2} & \ldots &
  z^{\mfirst,l}_{h+(\poolaout - 1)\times s,h+(\poolbout - 1)\times
  s,\pooldin}
\end{bmatrix} \in R^{h h \times \pooldin \poolaout \poolbout l}.
\end{equation}
We apply a max function to get the largest value of each column 
and its index in the range of $1,\ldots,h h$.
The resulting row vector has $\pooldin \poolaout \poolbout l$ elements;
see line \ref{list:p-pool|maxpool}. In line \ref{list:p-pool|Z-d-rows}, we
reformulate it to be
\begin{equation*}
\pooldout \times \poolaout \poolbout l
\end{equation*}
as the output $Z^{\oupu,i},\ \forall i$.

\par Next we find linear indices that correspond to the largest elements
obtained from \eqref{pool:matphiz}. Because of operations discussed in Section
III of supplementary materials, we decide to record
linear indices in each $Z^{\inpu, i}$ corresponding to the selected elements, rather than
linear indices in the whole matrix \eqref{impl:Z} of all $Z^{m,i},\ \forall i$. 
We begin with obtaining the following vector of linear indices of $Z^{\inpu,
i}$:
\begin{equation}
  \begin{bmatrix}
    1  \\ \vdots  \\ \pooldin \poolain \poolbin
  \end{bmatrix}.
  \label{Zconv-linearind}
\end{equation}
Then we generate
\begin{equation}
  \phi(\eqref{Zconv-linearind}),
  \label{pool:phiz-linearidx}
\end{equation}
which has $h h \pooldout \poolaout \poolbout$ elements; see line
\ref{list:p-pool|phi-poolidx}.
Next, we mentioned that in line \ref{list:p-pool|maxpool}, not only the maximal value in each
sub-region is obtained, but also the corresponding index in $\{
1,\ldots,h h \}$ is derived. Therefore, for the selected max values of all instances, their
positions in the range of
\begin{equation*}
  1,\ \ldots, h h \pooldout \poolaout \poolbout
\end{equation*}
are
\begin{equation}
  \label{pool:phiz-maxval-linearidx-norm}
  \mat\left(
  \begin{bmatrix}
    \text{row indices of}\\
    \text{max values in \eqref{pool:matphiz}}
  \end{bmatrix}\right)_{\pooldout \poolaout \poolbout \times l }
  + h h \left(
    \begin{bmatrix}
      0 \\ \vdots \\ \pooldout \poolaout \poolbout - 1
  \end{bmatrix} \otimes \mathds{1}_l^T
  \right);
\end{equation}
see line \ref{list:p-pool|phi-pool-linearidx}. Next in line
\ref{list:p-pool|poolidx} we use \eqref{pool:phiz-maxval-linearidx-norm} to
extract values in \eqref{pool:phiz-linearidx} and obtain linear indices of the 
selected max values in each $Z^{\inpu, i}$.
To be more precise, the resulting matrix is
\begin{equation}
  \begin{bmatrix}
  \begin{matrix}
    \text{Column index of non-zero} \\
    \text{in each row of } P^{m,1}_{\pool}
  \end{matrix} &
    \hdots &
  \begin{matrix}
    \text{Column index of non-zero} \\
    \text{in each row of } P^{m,l}_{\pool}
  \end{matrix}
  \end{bmatrix} \in R^{\pooldout \poolaout \poolbout \times l}.
  \label{idxpool}
\end{equation}
The reason is that because $P^{m,i}_{\pool}$ is a $0/1$
matrix and each row contains exactly only one value ``1'' to indicate the
selected entry by max pooling, we collects the column
index of the non-zero at each row to be a vector for future use.

\renewcommand{\baselinestretch}{1.3}
\begin{lstlisting}[caption = {{\sl MATLAB} implementation for $P^{m,i}_{\pool}$},
label={list:p-pool}, float=t]
function [Zout, idx_pool] = maxpooling(model, net, m)

a = model.ht_conv(m);
b = model.wd_conv(m);
d = model.ch_input(m+1);
h = model.wd_subimage_pool(m);

% Z input: sigma(S_m)
Z = net.Z{m+1};

P = net.idx_phiZ_pool{m};               %( \label{list:p-pool|P} %)
Z = reshape(Z, d*a*b, []);              %( \label{list:p-pool|gen-vec-phi-st} %)
Z = Z(P, :);                            %( \label{list:p-pool|gen-vec-phi-ed} %)
[Z, max_id] = max(reshape(Z, h*h, []));     %( \label{list:p-pool|maxpool} %)
Zout = reshape(Z, d, []);               %( \label{list:p-pool|Z-d-rows} %)

outa = model.ht_input(m+1);
outb = model.wd_input(m+1);
max_id = reshape(max_id, d*outa*outb, []) + h*h*[0:d*outa*outb-1]';  %( \label{list:p-pool|phi-pool-linearidx} %)

idx_pool = [1:d*a*b];                   %( \label{list:p-pool|Z-linear-idx} %)
idx_pool = idx_pool(P);                 %( \label{list:p-pool|phi-poolidx} %)
idx_pool = idx_pool(max_id);                %( \label{list:p-pool|poolidx} %)

\end{lstlisting}
\renewcommand{\baselinestretch}{2}

\subsection{Details of Padding Operation}
\label{subsec:impl-padding}
To implement zero-padding, we first capture the linear indices of the input image in the padded image.
For example, if the size of the input image is $3 \times 3$ and the output padded image is $5\times5$,
we have
\begin{equation*}
\begin{bmatrix}
0 & 0 & 0 & 0 & 0 \\
0 & 1 & 1 & 1 & 0 \\
0 & 1 & 1 & 1 & 0 \\
0 & 1 & 1 & 1 & 0 \\
0 & 0 & 0 & 0 & 0
\end{bmatrix},
\end{equation*}
where ``1'' values indicate positions of the input image. Based on the column-major order, we derive
\begin{align*}
  \pad_idx &= \{7,\ 8,\ 9,\ 11,\ 12,\ 13,\ 16,\ 17,\ 18\}.
\end{align*}
This index set, obtained in the beginning of the training procedure, is used in the
following situations. First, $\pad_idx$ contains row indices in
$P^{m}_{\padding}$ of \eqref{padding} that correspond to the input image. We can use
it to conduct the padding operation in \eqref{padding}. Second, from
\eqref{gd:dXidZ} and \eqref{jaco:zL-z} in gradient and Jacobian evaluations, we need
\begin{equation*}
  \bv^T P^{m}_{\padding}.
\end{equation*}
This can be considered as the inverse of the padding operation: we would like
to remove zeros and get back the original image. We give details of finding
$\pad_idx$ in Section II of supplementary materials.

\renewcommand{\baselinestretch}{1.3}
\begin{lstlisting}[caption = {{\sl MATLAB} implementation for the index of
zero-padding}, label={list:fun-zero-padding-idx}, float=t]
function [idx_pad] = find_index_padding(model,m)

a = model.ht_input(m);
b = model.wd_input(m);
p = model.wd_p(m);

newa = 2*p + a;
idx_pad = reshape( (p+1:p+a)' + newa*(p:p+b-1), [], 1);
\end{lstlisting}
\renewcommand{\baselinestretch}{2}

\subsection{Evaluation of $(\bv^i)^T P^{m}_{\phi}$ and $(\bv^i)^T
P^{m,i}_{\pool}$ in Gradient and Jacobian Evaluations}

We show that several operations in gradient and Jacobian evaluations are either 
$(\bv^i)^T P^{m}_{\phi}$ or $(\bv^i)^T P^{m,i}_{\pool}$, where $\bv^i$ is a
vector. They can be calculated by a similar program. We
give Listing \ref{list:vTP} to conduct these operations with details explained
below.

\subsubsection{Evaluation of $(\bv^i)^T P^{m}_{\phi}$}
\label{subsubsec:impl-vTP-phi}

For \eqref{gd:dXidZ} and \eqref{jaco:zL-z}, the following operation is applied.
\begin{equation}
\label{vTP}
(\bv^i)^T P^{m}_{\phi},
\end{equation}
where
\begin{equation*}
\bv^i = \vec\left((W^m)^T \frac{\partial \xi_i}{\partial S^{\mfirst,i}}\right)
\end{equation*}
for \eqref{gd:dXidZ} and
\begin{equation}
\bv^i_u = \vec\left((W^m)^T \frac{\partial z_u^{\mout,i}}{\partial
S^{\mfirst,i}}\right),\ u = 1,\ldots, n_\mout
\label{jaco:vTP-v}
\end{equation}
for \eqref{jaco:zL-z}.

\par Consider the same example in Section \ref{subsec:impl-phi}. We note that
\begin{equation}
\label{vTP-example}
(P^{m}_{\phi})^T \bv^i = \left[\ v_1\ \ v_2 + v_5\ \ v_6\ \ v_3\ \ v_4+v_7\ \ v_8\ \right]^T,
\end{equation}
which is a kind of ``inverse'' operation of $\phi(\padding(Z^{m,i}))$:
we accumulate elements in $\phi(\padding(Z^{m,i}))$ back to their original positions in
$\padding(Z^{m,i})$. In {\sl MATLAB}, given indices in \eqref{linearidx-eg}, a function
$\accumarray$ can directly generate the vector \eqref{vTP-example}.

\par To calculate \eqref{gd:dXidZ} over a batch of
instances, we aim to have
\begin{equation}
  \begin{bmatrix}
    (P^{m}_{\phi})^T \bv^1  \\
    \vdots  \\
    (P^{m}_{\phi})^T \bv^l  \\
  \end{bmatrix}^T.
  \label{PTv}
\end{equation}
We can manage to apply {\sl MATLAB}'s $\accumarray$ on the vector
\begin{equation}
  \begin{bmatrix}
    \bv^1  \\
    \vdots  \\
    \bv^l  \\
  \end{bmatrix},
  \label{vTP:accumarray-v}
\end{equation}
by giving the following indices as the input.
\begin{equation}
  \begin{bmatrix}
    \eqref{idxphi}  \\
    \eqref{idxphi} + \convain \convbin \convdin \mathds{1}_{h^m
    h^m \convdin \convaout \convbout}  \\
    \eqref{idxphi} + 2 \convain \convbin \convdin \mathds{1}_{h^m
    h^m \convdin \convaout \convbout}  \\
    \vdots  \\
    \eqref{idxphi} + (l-1) \convain \convbin \convdin
    \mathds{1}_{h^m h^m \convdin \convaout \convbout}  \\
  \end{bmatrix},
  \label{vTP:accumarray-idx}
\end{equation}
where from Section \ref{subsubsec:summary-convlayer},
\begin{align*}
  &\convain \convbin \convdin \text{ is the size of } \padding(Z^{\mfirst,i}), \text{ and}\\
  &h^m h^m \convdin \convaout \convbout \text{ is the size of } \phi(\padding(Z^{\mfirst,i})) \text{ and }
  \bv_i.
\end{align*}
That is, by using the offset $(i-1) \convain \convbin \convdin$, $\accumarray$
accumulates $\bv^i$ to the following positions:
\begin{equation}
  (i - 1) \convain \convbin \convdin + 1,\ \ldots,\ i \convain \convbin \convdin.
  \label{acuumarray:inst-offset}
\end{equation}
To obtain \eqref{vTP:accumarray-v}, we can do a matrix-matrix multiplication as
follows.
\begin{align}
  \eqref{vTP:accumarray-v}
  = \vec\left((W^m)^T \begin{bmatrix}
    \frac{\partial \xi_1}{\partial S^{\mfirst,1}}  &
      \hdots  &
    \frac{\partial \xi_l}{\partial S^{\mfirst,l}}
\end{bmatrix} \right).
\label{impl:WmdXidS}
\end{align}
From \eqref{impl:WmdXidS}, we can see why $\partial \xi_i
/ \partial \vec(S^{\mfirst,i})^T$ over a batch of instances are stored in the form of
\eqref{impl:dXidS}.
In line \ref{list:vTP|idx}, the indices shown in \eqref{vTP:accumarray-idx} are
generated and the variable ${\tt V(:)}$ in line \ref{list:vTP|v} is the vector
\eqref{vTP:accumarray-v} calculated by \eqref{impl:WmdXidS}.

To calculate \eqref{jaco:zL-z} over a batch of
instances, similar to \eqref{PTv} we conduct
\begin{equation}
  \begin{bmatrix}
    (P^{m}_{\phi})^T \bv^1_1  \\
    \vdots  \\
    (P^{m}_{\phi})^T \bv^1_{n_\mout}  \\
    \vdots  \\
    (P^{m}_{\phi})^T \bv^l_{n_\mout}  \\
  \end{bmatrix}^T,
  \label{jaco:PTv}
\end{equation}
where
\begin{equation*}
  \bv^i_u = \vec \left( (W^m)^T \partialdiff{z^{\mout,i}_u}{S^{\mfirst,i}} \right) \in
  R^{h^m h^m \convdin \convaout \convbout \times 1},\ u = 1, \hdots, n_\mout.
\end{equation*}
Similar to \eqref{impl:WmdXidS}, we can calculate the vector
\begin{equation}
  \begin{bmatrix}
    \bv^1_1  \\
    \vdots  \\
    \bv^1_{n_\mout}  \\
    \vdots  \\
    \bv^l_{n_\mout}
  \end{bmatrix}
  \label{impl:dzdZ_phi}
\end{equation}
  by
\begin{equation}
  \vec\left((W^m)^T \begin{bmatrix}
      \frac{\partial z^{\mout,1}_1}{\partial S^{\mfirst,1}}  &
      \hdots  &
      \frac{\partial z^{\mout,1}_{n_\mout}}{\partial S^{\mfirst,1}}  &
      \hdots  &
      \frac{\partial z^{\mout,l}_{n_\mout}}{\partial S^{\mfirst,l}}
\end{bmatrix} \right).
\label{impl:WmdzdS}
\end{equation}
The formulation in \eqref{impl:WmdzdS} leads us to store
\begin{equation*}
  \partialdiff{\bz^{\mout, i}}{\vec(S^{m,i})},\ \forall i
\end{equation*}
in the form of \eqref{impl:dzdS}.

From \eqref{jaco:PTv}, because each vector $\bv^i_u$ is accumulated to the
following positions:
\begin{equation*}
  ((i - 1) n_\mout + (u - 1)) \convain \convbin \convdin + 1,\ \hdots,\
  ((i - 1) n_\mout + u) \convain \convbin \convdin,
\end{equation*}
we can apply $\accumarray$ on the vector \eqref{impl:dzdZ_phi}
with the following input indices.
\begin{equation}
  \begin{bmatrix}
    \eqref{idxphi}  \\
    \vdots  \\
    \eqref{idxphi} + (n_\mout-1) \convdin \convain \convbin
    \mathds{1}_{h^m h^m \convdin \convaout \convbout}  \\
    \vdots  \\
    \eqref{idxphi} + (n_\mout l - 1) \convdin \convain \convbin
    \mathds{1}_{h^m h^m \convdin \convaout \convbout}  \\
  \end{bmatrix}.
  \label{vTP:Jacobian-idx}
\end{equation}
The implementation is the same as that for evaluating \eqref{gd:dXidZ}, except
that \eqref{jaco:PTv} involves $n_{L+1} l$ vectors rather than $l$.

\subsubsection{Evaluation of $(\bv^i)^T P^{m,i}_{\pool}$}
\label{subsubsec:impl-vTP-pool-gradient-Jacobian}

We discuss details in supplementary materials.

\renewcommand{\baselinestretch}{1.3}
\begin{lstlisting}[caption = {{\sl MATLAB} implementation to evaluate $(\bv^i)^T
P^{m}_{\phi}$ and $(\bv^i)^T P^{m,i}_{\pool}$}, float=t, label={list:vTP}]
function vTP = vTP(param, model, net, m, V, op)
% output vTP: a row vector, where mat(vTP) is with dimension $d_prev a_prev b_prev \times num_v$.

nL = param.nL;
num_data = net.num_sampled_data;

switch op
case {'pool_gradient', 'pool_Jacobian'}
	a_prev = model.ht_conv(m);
	b_prev = model.wd_conv(m);
	d_prev = model.ch_input(m+1);
	if strcmp(op, 'pool_gradient')
		num_v = num_data;
		idx = net.idx_pool{m} + [0:num_data-1]*d_prev*a_prev*b_prev;  %( \label{list:vTP|pool-gradient} %)
	else
		num_v = nL*num_data;
		idx = reshape(net.idx_pool{m}, [], 1, num_data) + reshape([0:nL*num_data-1]*d_prev*a_prev*b_prev, 1, nL, num_data);  %( \label{list:vTP|pool-Jacobian} %)
	end
case {'phi_gradient', 'phi_Jacobian'}
	a_prev = model.ht_pad(m);
	b_prev = model.wd_pad(m);
	d_prev = model.ch_input(m);

	if strcmp(op, 'phi_gradient'); num_v = num_data; else; num_v = nL*num_data; end

	idx = net.idx_phiZ{m}(:) + [0:num_v-1]*d_prev*a_prev*b_prev; %( \label{list:vTP|idx} %)
otherwise
	error('Unknown operation in function vTP.');
end

vTP = accumarray(idx(:), V(:), [d_prev*a_prev*b_prev*num_v 1])'; %( \label{list:vTP|v} %)

\end{lstlisting}
\renewcommand{\baselinestretch}{2}

\subsection{Gauss-Newton Matrix-Vector Products}
\label{subsec:impl-GN-matrix-vector}

\par To derive \eqref{Gv:block-diag}, we first calculate
\begin{equation}
\begin{bmatrix}
\sum_{m=1}^L J^{m,1} \bv^m \\ \vdots \\ \sum_{m=1}^L J^{m,l} \bv^m 
\end{bmatrix} \in R^{n_\mout l \times 1}.
\label{matlab-Jv}
\end{equation}
From \eqref{Jv}, for a particular $m$, we have
\begin{align}
\begin{bmatrix}
J^{m,1} \bv^m \\ \vdots \\ J^{m,l} \bv^m 
\end{bmatrix}
&=
\begin{bmatrix}
\frac{\partial \bz^{\mout,1}}{\partial \vec(S^{\mfirst,1})^T} \vec\left(
  \mat(\bv^m) \begin{bmatrix} \phi(\padding(Z^{\mfirst,1})) \\  \mathds{1}^T_{\convaout
\convbout} \end{bmatrix}\right)
\\
\vdots
\\
\frac{\partial \bz^{\mout,l}}{\partial \vec(S^{\mfirst,l})^T} \vec\left(
  \mat(\bv^m) \begin{bmatrix} \phi(\padding(Z^{\mfirst,l})) \\  \mathds{1}^T_{\convaout
\convbout} \end{bmatrix}\right)
\end{bmatrix}  \nonumber\\
&=
\begin{bmatrix}
\frac{\partial \bz^{\mout,1}}{\partial \vec(S^{\mfirst,1})^T} \bp^{m,1}
\\
\vdots
\\
\frac{\partial \bz^{\mout,l}}{\partial \vec(S^{\mfirst,l})^T} \bp^{m,l}
\end{bmatrix},
\label{Jv:Jp}
\end{align}
where
\begin{equation*}
\mat(\bv^m) \in R^{d^\msec \times (h^m h^m d^{\mfirst} + 1)} 
\end{equation*}
and
\begin{equation}
  \bp^{m,i}
  = \vec \left(
    \mat(\bv^m)
    \begin{bmatrix}
      \phi(\padding(Z^{\mfirst,i})) \\
      \mathds{1}^T_{\convaout \convbout}
    \end{bmatrix}
  \right). 
  \label{Jv:p_i}
\end{equation}

A {\sl MATLAB} implementation for \eqref{matlab-Jv} is shown in Listing \ref{list:Jv}.
Given $\bv^m$, we calculate
\begin{equation}
  \mat(\bv^m)
  \begin{bmatrix}
    \phi(\padding(Z^{\mfirst,1})) & \cdots & \phi(\padding(Z^{\mfirst,l})) \\
    \mathds{1}^T_{\convaout \convbout} & \cdots &  \mathds{1}^T_{\convaout
  \convbout}
  \end{bmatrix} \in R^{\convdout \times \convaout \convbout l};
  \label{Jv:v_phiZ}
\end{equation}
see line \ref{list:Jv|p}. Next, we calculate
\begin{equation}
  J^{m,i} \bv^m
  = \frac{\partial \bz^{\mout,i}}{\partial \vec(S^{\mfirst,i})^T} \bp^{m,i},\ i=1,\ldots,l.
  \label{impl:Jv}
\end{equation}
Because \eqref{impl:Jv} involves $l$ independent matrix-vector products, we consider
the following trick to avoid a {\tt for} loop in a {\sl MATLAB} script.
We note that \eqref{impl:Jv} can be calculated by summing up all rows of the following matrix 
\begin{equation}
\left[ \frac{\partial z^{\mout,i}_1}{\partial \vec(S^{\mfirst,i})} \cdots \frac{\partial z^{\mout,i}_{n_{L+1}}}{\partial \vec(S^{\mfirst,i})} 
\right]_{d^{m+1} a^m_{\conv} b^m_{\conv} \times n_{L+1}}
\odot 
\left[\bp^{m,i} \cdots \bp^{m,i}\right]_{d^{m+1} a^m_{\conv} b^m_{\conv} \times n_{L+1}}.
\label{ZS-odot-p}
\end{equation}
The result will be a row vector of $1 \times n_{L+1}$, which is the transpose of $J^{m,i} \bv^m$.
To do the above operation on all instances together, we reformulate \eqref{impl:dzdS} and \eqref{Jv:v_phiZ}
respectively to the following three-dimensional matrices:
\begin{equation*}
  \convdout \convaout \convbout \times n_\mout \times l
  \quad\text{and}\quad \convdout \convaout \convbout \times 1 \times l.
\end{equation*}
We then apply the {\tt .*} operator in {\sl MATLAB} and sum results along
the first dimension; see line
\ref{list:Jv|Jp}. The resulting matrix has the size
\begin{equation*}
  1 \times n_\mout \times l
\end{equation*}
and can be aggregated to the vector in \eqref{Jv:Jp}; see line
\ref{list:Jv|sumJv}.

\renewcommand{\baselinestretch}{1.3}
\begin{lstlisting}[caption = {{\sl MATLAB} implementation for $J\bv$}, label={list:Jv}, float=t]
function Jv = Jv(param, model, net, v)

nL = param.nL;
L = param.L;
LC = param.LC;
num_data = net.num_sampled_data;
var_ptr = model.var_ptr;
Jv = zeros(nL*num_data, 1);

for m = L : -1 : LC+1
	var_range = var_ptr(m) : var_ptr(m+1) - 1;
	n_m = model.full_neurons(m-LC);

	p = reshape(v(var_range), n_m, []) * [net.Z{m}; ones(1, num_data)];
	p = sum(reshape(net.dzdS{m}, n_m, nL, []) .* reshape(p, n_m, 1, []),1);
	Jv = Jv + p(:);
end

for m = LC : -1 : 1
	var_range = var_ptr(m) : var_ptr(m+1) - 1;
	ab = model.ht_conv(m)*model.wd_conv(m);
	d = model.ch_input(m+1);

	p = reshape(v(var_range), d, []) * [padding_and_phiZ(model, net, m); ones(1, ab*num_data)];  %( \label{list:Jv|p} %)
	p = sum(reshape(net.dzdS{m}, d*ab, nL, []) .* reshape(p, d*ab, 1, []),1);  %( \label{list:Jv|Jp} %)
	Jv = Jv + p(:);  %( \label{list:Jv|sumJv} %)
end
\end{lstlisting}
\renewcommand{\baselinestretch}{2}

\par 
After deriving \eqref{matlab-Jv}, from \eqref{BJv-inst}, we must calculate
\begin{equation}
  \bq
  = \begin{bmatrix}
    \bq^1 \\
    \vdots \\
    \bq^{l}
  \end{bmatrix}
  = \begin{bmatrix}
    B^1 \sum_{m=1}^L J^{m,1} \bv^m \\ \vdots \\ B^l \sum_{m=1}^L J^{m,l} \bv^m
  \end{bmatrix}.
  \label{matlab-BJv}
\end{equation}
From \eqref{B-def}, \eqref{matlab-BJv} can be derived by multiplying every element of \eqref{matlab-Jv} by two.

Next, for each layer $m$, from \eqref{Gv:block-diag} and \eqref{JtV} we calculate
\begin{align}
  &\sum_{i=1}^l J^{m,i} \bq^i  \\
=\ &\sum_{i=1}^l \vec\left( \mat\left(\left(\frac{\partial \bz^{\mout,i}}{\partial
\vec(S^{\mfirst,i})^T}\right)^T\bq^{i}\right)_{\convdout \times \convaout \convbout}
\left[\phi(\padding(Z^{\mfirst,i}))^T\ \mathds{1}_{\convaout \convbout}\right]\right)\nonumber\\
=\ &
\vec\left(
\begin{bmatrix}
\mat(\bu^{m,1})_{\convdout \times \convaout \convbout} & \hdots &
  \mat(\bu^{m,l})_{\convdout \times \convaout \convbout}
\end{bmatrix}
\begin{bmatrix}
\phi(\padding(Z^{\mfirst,1}))^T & \mathds{1}_{\convaout \convbout} \\
\vdots & \vdots \\
\phi(\padding(Z^{\mfirst,l}))^T & \mathds{1}_{\convaout \convbout}
\end{bmatrix}\right),
\label{matlab-JTP}
\end{align}
where
\begin{equation*}
  \bu^{m,i}
  = \left( \frac{\partial \bz^{\mout,i}}{\partial \vec(S^{\mfirst,i})^T} \right)^T
  \bq^{i}.
\end{equation*}

\par A {\sl MATLAB} implementation for \eqref{matlab-JTP} is shown in Listing \ref{list:JTq}.
To begin, we have the matrix \eqref{impl:dzdS} and the vector $\bq$ in \eqref{matlab-BJv}. We
reshape \eqref{impl:dzdS} to
\begin{equation}
  \begin{bmatrix}
    \partialdiff{z^{\mout,1}_{1}}{\vec(S^{\mfirst, 1})} & \hdots &
    \partialdiff{z^{\mout,1}_{n_\mout}}{\vec(S^{\mfirst, 1})} & \hdots &
    \partialdiff{z^{\mout,l}_{n_\mout}}{\vec(S^{\mfirst, l})}
  \end{bmatrix}
  \in R^{\convdout \convaout \convbout \times n_\mout l}.
  \label{impl:dzdS-dab-nLl}
\end{equation}
Then we calculate
\begin{equation*}
  \begin{bmatrix}
    \bu^{m,1} & \ldots & \bu^{m,l}
  \end{bmatrix}
\end{equation*}
together by reshaping
\begin{equation}
  \begin{bmatrix}
    \eqref{impl:dzdS-dab-nLl}
  \end{bmatrix}
  \odot \left(
  \mathds{1}_{\convdout \convaout \convbout} \bq^T \right) \in R^{\convdout \convaout
  \convbout \times n_\mout l}
  \label{impl:JTq}
\end{equation}
to
\begin{equation*}
  R^{\convdout \convaout \convbout \times n_\mout \times l}
\end{equation*}
and summing along the second dimension;
see line \ref{list:JTq|r-st}-\ref{list:JTq|r-ed}.
Finally, we calculate \eqref{matlab-JTP} in line \ref{list:JTq|JTq_m}.
\renewcommand{\baselinestretch}{1.3}
\begin{lstlisting}[caption = {{\sl MATLAB} implementation for $J^T\bq$}, label={list:JTq}, float=t]
function u = JTq(param, model, net, q)

nL = param.nL;
L = param.L;
LC = param.LC;
num_data = net.num_sampled_data;
var_ptr = model.var_ptr;
n = var_ptr(end) - 1;
u = zeros(n, 1);

for m = L : -1 : LC+1
	var_range = var_ptr(m) : var_ptr(m+1) - 1;

	u_m = net.dzdS{m} .* q';
	u_m = sum(reshape(u_m, [], nL, num_data), 2);
	u_m = reshape(u_m, [], num_data) * [net.Z{m}' ones(num_data, 1)];
	u(var_range) = u_m(:);
end

for m = LC : -1 : 1
	a = model.ht_conv(m);
	b = model.wd_conv(m);
	d = model.ch_input(m+1);
	var_range = var_ptr(m) : var_ptr(m+1) - 1;

	u_m = reshape(net.dzdS{m}, [], nL*num_data) .* q';   %( \label{list:JTq|r-st} %)
	u_m = sum(reshape(u_m, [], nL, num_data), 2);      %( \label{list:JTq|r-ed} %)
	u_m = reshape(u_m, d, []) * [padding_and_phiZ(model, net, m)' ones(a*b*num_data, 1)];  %( \label{list:JTq|JTq_m} %)
	u(var_range) = u_m(:);
end
\end{lstlisting}
\renewcommand{\baselinestretch}{2}

\subsection{Mini-Batch Function and Gradient Evaluation}
\label{subsec:mini-batch-Newton}
Later in Section \ref{subsec:analyze-memory} we discuss details of memory usage,
where one important conclusion is that in several places of the Newton method, the
memory consumption is proportional to the number of data. This fact causes
difficulties in handling large data sets, so here we discuss some
implementation techniques to reduce the memory usage.

\par In the subsampled Newton method discussed in Section
\ref{subsec:proposed-alg}, a subset $S$ of the training data is used to derive
the subsampled Gauss-Newton matrix for approximating the Hessian matrix.
While a motivation of this technique is to trade a slightly less accurate direction for shorter running time per iteration,
it also effectively reduces the memory consumption.
For example, at the $m$th convolutional layer, we only need to store the following matrices
\begin{equation}
\frac{\partial \bz^{\mout,i}}{\partial \vec(S^{m,i})^T},\ \forall i \in S
\label{impl:mini-batch-dz-ds}
\end{equation}
for the Gauss-Newton matrix-vector products.

However, in function and gradient evaluations we still need the whole training data.
Fortunately, both operations involve the summation of independent results over
all instances, so we follow
\cite{CCW16a} to have a mini-batch setting. By splitting the index set
$\{1,\ldots,l\}$ of data to, for example, $R$ equal-sized subsets
$S_1,\ldots,S_R$, we sequentially calculate the result corresponding to each
subset and accumulate them for the final output.
Take the function evaluation as an example. For each subset, we must store only
\begin{equation*}
Z^{m,i},\ \forall m,\ \forall i \in S_r,
\end{equation*}
so the memory usage can be dramatically reduced.

\par For the Gauss-Newton matrix-vector product, to calculate
\eqref{impl:mini-batch-dz-ds} under the subsampled scheme, we have a set $S$ and
use $Z^{m,i},\ \forall i \in S$. However, under the mini-batch setting,
the needed values may not be kept in the process of function evaluations. Our
strategy is to let the last subset $S_R$ be the same subset used for the
sub-sampled Hessian. Then we can preserve the needed $Z^{m,i}$ for Gauss-Newton
matrix-vector products.

\section{Analysis of Newton Methods for CNN}
\label{sec:analysis}
In this section, based on the implementation details in Section \ref{sec:impl}, we analyze the memory and computational
cost per iteration. We consider that all training instances are used. 
If the subsampled Hessian in Section \ref{sec:HessianFree} is considered, 
then in the Jacobian calculation and the Gauss-Newton matrix vector products, the number of instances $l$
should be replaced by the subset size $|S|$.
Furthermore, if mini-batch function and gradient evaluation in Section \ref{subsec:mini-batch-Newton} is applied, the number of instance $l$
in the function and gradient evaluation can also be replaced by
$|S|$.\footnote{We mentioned in Section \ref{subsec:mini-batch-Newton} that for simplicity,
we set the mini-batch size be $|S|$ in order to cooperate with the subsampled Hessian Newton methods.}

\par In this discussion we exclude the padding and the pooling operations
because first they are optional steps and second they are not the bottleneck.
In addition, for simplicity, the bias term is not considered.
\subsection{Memory Requirement}
\label{subsec:analyze-memory}
\begin{enumerate}[(1)]
  \item Weight matrix:
    For every layer, we must store
    \begin{equation*}
      W^m,\ m = 1,\ldots,L.
    \end{equation*}
    From \eqref{conv-wmatrix} and \eqref{fc-wmatrix}, the memory usage is
    \begin{equation*}
      \sum_{m=1}^{L^c} \left(d^\msec h^m h^m d^{\mfirst}\right) +
      \sum_{m=L^c+1}^{L} \left(n_{\msec} n_{\mfirst}\right).
    \end{equation*}
  \item Gradient vector:
	For \eqref{gradient-w}, the following matrix must be stored.
    \begin{equation*}
      \frac{\partial f}{\partial \vec(W^m)^T},\ m=1,\ldots,L.
    \end{equation*}
    Therefore, the memory usage is
    \begin{equation*}
      \sum_{m=1}^{L^c} \left(d^\msec h^mh^md^{\mfirst}\right) + \sum_{m=L^c+1}^L
      \left(n_\msec n_{\mfirst}\right).
    \end{equation*}
  \item $P^{m}_{\phi}$:
    In Section \ref{subsec:impl-phi}, we store
    each position's corresponding linear index in $Z^{\mfirst,i}$ in order to
    construct $\phi(Z^{\mfirst,i})$. 
	The memory usage is
    \begin{equation*}
      \sum_{m=1,\ldots,L^c} \left( h^m h^m \convdin \convaout \convbout \right).
    \end{equation*}
  \item Function evaluation:
    From Section \ref{sec:FunEval}, we store
    \begin{equation*}
      Z^{\mfirst,i},\ m = 1,\ldots,L+1,\ \forall i.
    \end{equation*}
    Therefore, the memory usage is
    \begin{equation}
      l \times \left(\sum_{m=1}^{L^c} d^m a^m b^m + \sum_{m=L^c + 1}^{L+1} n_m \right).
	\label{memory:z}
    \end{equation}
  \item $\phi(\padding(Z^{\mfirst,i}))$:
	From \eqref{conv-f-ztos}, \eqref{conv-dxidw}, \eqref{Jv}, and \eqref{JtV},
  $\phi(\padding(Z^{\mfirst,i}))$ is temporarily needed and its memory usage is
	\begin{equation*}
		l \times \max_{m=1,\ldots,L^c} \left( h^m h^m \convdin \convaout \convbout \right). 
	\end{equation*}
	Comparing to \eqref{memory:z}, the temporary memory for $\phi(\padding(Z^{\mfirst,i}))$ is insignificant. 
  \item Gradient evaluation:
    To obtain the gradient in each layer $m$, we also need to have the matrix
    \begin{equation*}
      \frac{\partial \xi_i}{\partial \vec(S^{m,i})^T}
    \end{equation*}
    for calculating
    \begin{equation*}
      \frac{\partial \xi_i}{\partial \vec(S^{m-1,i})^T}
    \end{equation*}
    in the backward process.
	Note that there is no need to keep the matrices of all layers. All we have to
	store is the matrices for two adjacent layers. Thus, the memory
    usage is
    \begin{align*}
      l \times \max_{m=1,\ldots,L^c}\left(d^{\mfirst} a^{\mfirst}_{\conv}
        b^{\mfirst}_{\conv} +
      d^\msec a^\msec_{\conv} b^\msec_{\conv}\right)
    \end{align*}
    for the convolutional layers and
    \begin{align*}
      l \times \max_{m=L^c+1,\ldots,L} \left(n_{\mfirst} + n_\msec\right).
    \end{align*}
    for the fully-connected layers. This is much smaller than \eqref{memory:z}.
  \item Jacobian evaluation and Gauss-Newton matrix-vector products:
    Besides $W^m$ and $Z^{\mfirst,i}$, from \eqref{Jv}, \eqref{JtV},
    \eqref{full-Jv} and \eqref{full-JtV}, we
      explicitly store
      \begin{equation*}
        \frac{\partial \bz^{\mout,i}}{\partial \vec(S^{m,i})^T},\
        m=1,\ldots,L,\ \forall i.
      \end{equation*}
      Thus, the memory usage is\footnote{Note that the dimension
        of $\bs^{m,i}$ is $R^{n_{m+1}}$.}.
      \begin{equation}
        l \times n_\mout \times \left( \sum^{L^c}_{m=1} \convdout \convaout
          \convbout + \sum^{L}_{m=L^c+1} n_{m+1}\right).
        \label{memory:J}
      \end{equation}
\end{enumerate}

\par From the above discussion, \eqref{memory:z} dominate the memory usage in the function and Gradient evaluation. On the other hand,
\eqref{memory:J} is the main cost for the Jacobian evaluation and Gauss-Newton matrix-vector products.
The bottleneck is \eqref{memory:J} because it is $n_\mout$ times more than \eqref{memory:z}.
To reduce the memory consumption, as mentioned, the sub-sampled Hessian technique in Section \ref{sec:HessianFree} reduces the usage in \eqref{memory:J},
while for \eqref{memory:z} we use the mini-batch function and gradient evaluation in Section \ref{subsec:mini-batch-Newton}.

\subsection{Computational Cost}
\label{subsec:analyze-comp-cost}
We show the computational cost for the $m$th convolutional/fully-connected
layer.
\begin{enumerate}[(1)]
\item
Function evaluation:
\begin{itemize}
\item
Convolutional layers:
From \eqref{fun-phi}, \eqref{conv-f-ztos}, and \eqref{conv-f-stoz}, the computational cost is
\begin{equation*}
  \mathcal {O} (l\times h^m h^m \convdin \convdout \convaout \convbout).
\end{equation*}
\item
Fully-connected layers:
From \eqref{full-f-s} and \eqref{full-f-stoz}, the computational cost is
\begin{equation*}
\mathcal {O} (l \times n_\msec n_{\mfirst})
\end{equation*}
\end{itemize}
\item
Gradient evaluation:
\begin{itemize}
\item
Convolutional layers:
For \eqref{conv-dxidw}, the computational cost is
\begin{equation*}
  \mathcal {O} (l\times h^m h^m \convdin \convdout \convaout \convbout).
\end{equation*}
For \eqref{gd:xi-z-s}, the computational cost is
\begin{equation*}
  \mathcal {O}(l \times d^{\mfirst} a^{\mfirst} b^{\mfirst} ). 
\end{equation*}
For \eqref{gd:dXidZ}, the computational cost is
\begin{equation*}
  \mathcal {O} (l\times h^m h^m \convdin \convdout \convaout \convbout).
\end{equation*}
Therefore, the total computational cost for the gradient evaluation is
\begin{equation*}
  \mathcal {O} (l\times h^m h^m \convdin \convdout \convaout \convbout).
\end{equation*}
\item
Fully-connected layers:
From \eqref{gd:full-w} and \eqref{gd:full-z}, the computational cost is
\begin{equation*}
\mathcal {O} (l \times n_\msec n_{\mfirst}).
\end{equation*}
For \eqref{gd:full-s}, the cost is smaller.
Therefore, the total computational cost is 
\begin{equation*}
\mathcal {O} (l \times n_\msec n_{\mfirst}).
\end{equation*}
\end{itemize}
\item
Jacobian evaluation:
  \begin{itemize}
  \item
    Convolutional layers:
    The main computational cost is from \eqref{jaco:zL-z}:
    \begin{equation*}
      \mathcal {O} \left( l \times n_\mout \times h^m h^m \convdin \convdout
        \convaout \convbout \right),
    \end{equation*}
	while others are less significant.
  \item
    Fully-connected layers:
    From \eqref{jaco:xi-z-full}, the computational cost is
    \begin{equation*}
      \mathcal {O} (l \times n_\mout \times n_\msec n_{\mfirst}).
    \end{equation*}
\end{itemize}
\item
  CG: The computational cost is the number of CG iterations (\#CG) times the
  cost of  a Gauss-Newton matrix-vector
  product.
  \begin{itemize}
    \item
      Convolutional layers:
      The main computational cost is from \eqref{Jv} and \eqref{JtV}:      
	  \begin{equation*}
      \mathcal{O} \left( \text{\#CG} \times l \times \convdout h^m h^m \convdin
        \convaout \convbout \right),
      \end{equation*}
	  while the cost of \eqref{B-def} is insignificant. 
  \item
    Fully-connected layers:
    Similarly, the main computational cost is from \eqref{full-Jv} and \eqref{full-JtV}:
      \begin{equation*}
        \mathcal{O} \left( \text{\#CG} \times l \times n_\msec n_{\mfirst} \right).
      \end{equation*}
  \end{itemize}
\item
  line search: The computational cost is on multiple function evaluations.
  \begin{itemize}
  \item
      Convolutional layers:
      \begin{equation*}
        \mathcal{O} \left( \text{\#line search} \times l \times \convdout h^m h^m
          \convdin \convaout \convbout \right).
      \end{equation*}
  \item
    Fully-connected layers:
	\begin{equation*}
	\mathcal {O} ( \text{\#line search} \times l \times n_\msec n_{\mfirst}).
	\end{equation*}
  \end{itemize}
\end{enumerate}

We summarize the cost in a convolutional layer. Clearly, the cost is proportional to the number of instances, $l$.
After omitting the term $\mathcal{O} \left( h^m h^m \convdin \convdout \convaout
  \convbout \right)$ in all operations,
their cost can be compared in the following way.
\begin{equation*}
  \underbrace{l}_{\text{function/gradient}} \qquad \underbrace{l \times n_\mout}_{\text{Jacobian}} \qquad \underbrace{\text{\#CG} \times l}_{\text{CG}} \qquad \underbrace{\text{\#line search} \times l}_{\text{line search}}.
\end{equation*}
In general, the number of line search steps is small, so the CG procedure is often the bottleneck.
However, if the sub-sampled Hessian Newton method is applied, $l$ is replaced by the
size of the subset, $|S|$, for the cost in the Jacobian evaluation and CG.
Then the bottleneck may be shifted to function/gradient evaluations.
\par The discussion for the fully-connected layers is omitted because the result is similar to the convolutional layers.

\section{Experiments}
\label{sec:exps}
We choose the following image data sets for experiments.
All the data sets are publicly available\footnote{See \url{https://www.csie.ntu.edu.tw/~cjlin/libsvmtools/datasets/}.} and
the summary is in Table \ref{table:exps-CNN}.
\begin{itemize}
\item
{\sf MNIST:} This data set, containing hand-written digits, is a widely used benchmark for data classification \citep{YL98a}.
\item
{\sf SVHN:} This data set consists of the colored images of house numbers \citep{YN11a}.
\item
{\sf CIFAR10:} This data set, containing colored images, is a commonly used classification benchmark \citep{AK09a}.
\item
{\sf smallNORB:} This data set is built for 3D object recognition \citep{YL04b}.
The original dimension is $96 \times 96 \times 2$ because 
every object is taken two $96 \times 96$ grayscale images from the different
angles. These two images are then placed in two channels.
To reduce the training time, we downsample each channel of every object with the max pooling ($h=3, s = 3$) to the dimension $32 \times 32$.
\end{itemize}
\begin{table}[t]
\caption{Summary of the data sets, where $a^1 \times b^1 \times d^1$ represents the (height, width, channel)
of the input image, $l$ is the number of training data, $l_t$ is the number of
test data, and $n_\mout$ is the number of classes.}
\label{table:exps-CNN}
\begin{center}
\begin{tabular}{>{\centering\arraybackslash}p{2cm} >{\centering\arraybackslash}p{3cm} >{\centering\arraybackslash}p{1.5cm} >{\centering\arraybackslash}p{1.5cm} >{\centering\arraybackslash}p{1.5cm}}
	Data set & $a^1 \times b^1 \times d^1$ & $l$ & $l_t$ & $n_\mout$\\
\hline
	{\sf MNIST} & $28 \times 28 \times 1$ & $60,000$ & $10,000$ & $10$\\
\hline
	{\sf SVHN} & $32 \times 32 \times 3$ & $73,257$ & $26,032$ & $10$\\
\hline
	{\sf CIFAR10} & $32 \times 32 \times 3$ & $50,000$ & $10,000$ & $10$\\
\hline
	{\sf smallNORB} & $32 \times 32 \times 2$ & $24,300$ & $24,300$ & $5$\\
\end{tabular}
\end{center}
\end{table}

\par All the data sets were pre-processed by the following procedure.
\begin{enumerate}[(1)]
\item
Min-max normalization. That is, for each pixel of every image $Z^{1,i}$, we have
\begin{equation*}
	Z^{1,i}_{a,b,d} \leftarrow \frac{Z^{1,i}_{a,b,d} - \min}{\max - \min},
\end{equation*}
where $\max$/$\min$ is the maximum/minimum value of all pixels in $Z^{1,i}$.
\item
Zero-centering.
This is commonly applied before training CNN \citep{AK12b, MDZ14a}.
That is, for every pixel in image $Z^{1,i}$, we have
\begin{equation*}
Z^{1,i}_{a,b,d} \leftarrow Z^{1,i}_{a,b,d} - \mean(Z^{1,:}_{a,b,d}),
\end{equation*}
where $\mean(Z^{1,:}_{a,b,d})$ is the per-pixel mean value across all the training images.
\end{enumerate}

\par We consider two simple CNN structures shown in Table \ref{table:struc-CNN}.
The parameters used in our algorithm are given as follows. 
For the initialization, we follow \cite{KH15a} to set the weight values by
multiplying random values from the $\mathcal{N}(0,1)$ distribution and
\begin{equation*}
\sqrt{\frac{2}{n^m_{\text{in}}}},
\end{equation*}
where
\begin{equation*}
n^m_{\text{in}} = 
\begin{cases}
d^{\mfirst} \times h^m \times h^m & \text{if } m \leq L^c, \\
n_{\mfirst} & \text{otherwise.}
\end{cases}
\end{equation*}
For a CG procedure, it is terminated if the following relative stopping
condition holds or the number of CG
iterations reaches a maximal number of iterations (denoted as
$\text{CG}_{\text{max}}$).
\begin{equation}
||(G + \lambda \mathcal I)\bd + \nabla f(\btheta)|| \leq \sigma || \nabla f(\btheta) ||,
\label{cg-stopping}
\end{equation}
where $\sigma = 0.1$ and CG$_{\max} = 250$.
For the implementation of the Levenberg-Marquardt method, we set the initial $\lambda_1 = 1$ and (drop, boost, $\rho_{\text{upper}}$, $\rho_{\text{lower}}$) constants in \eqref{LM-rules} are ($2/3$, $3/2$, $0.75$, $0.25$).
In addition, the sampling rate for the Gauss-Newton matrix is set to $5\%$. The
value of $C$ in \eqref{L2loss} is set to $0.01l$.
We terminate the Newton method
after $100$ iterations.

\begin{table}[htp]
\caption{Structure of convolutional neural networks. ``conv'' indicates a
convolutional layer, ``pool'' indicates a pooling layer, and ``full'' indicates
a fully-connected layer.}
\label{table:struc-CNN}
\begin{center}
\begin{tabular}{>{\centering\arraybackslash}p{1.5cm} >{\centering\arraybackslash}p{3cm} >{\centering\arraybackslash}p{1.5cm} >{\centering\arraybackslash}p{1.5cm} >{\centering\arraybackslash}p{3cm} >{\centering\arraybackslash}p{1.5cm} >{\centering\arraybackslash}p{1.5cm}}
	\multicolumn{1}{c}{ } & \multicolumn{3}{c}{3-layer CNN} & \multicolumn{3}{c}{5-layer CNN} \\
\hline
	& filter size & $\#$filters & stride & filter size & $\#$filters & stride \\
\hline
	conv $1$ & $5 \times 5 \times 3$ & $32$ & $1$ & $5 \times 5 \times 3$ & $32$ & $1$ \\
\hline
	pool $1$ & $2 \times 2$ & - & $2$ & $2 \times 2$ & - & $2$\\
\hline
	conv $2$ & $3 \times 3 \times 32$ & $64$ & $1$ & $3 \times 3 \times 32$ & $32$ & $1$\\
\hline
	pool $2$ & $2 \times 2$ & - & $2$ & - & - & -\\
\hline
	conv $3$ & $3 \times 3 \times 32$ & $64$ & $1$ & $3 \times 3 \times 32$ & $64$ & $1$\\
\hline
	pool $3$ & $2 \times 2$ & - & $2$ & $2 \times 2$ & - & $2$\\
\hline
	conv $4$ &- &- & -& $3 \times 3 \times 64$ & $64$ & $1$\\
\hline
	pool $4$ & - & - & - & - & - & -\\
\hline
	conv $5$ &- &- &- & $3 \times 3 \times 64$ & $128$ & $1$\\
\hline
	pool $5$ &- &- &- & $2 \times 2$ & - & $2$\\
\hline
	full $1$ &- &- &- & - & - & - \\
\end{tabular}
\end{center}
\end{table}

\subsection{Preliminary Comparison Between Newton and Stochastic Gradient Methods}
\label{subsec:exps-compare}
The goal is to compare SG methods with the proposed subsampled Newton method for CNN.
For SG methods, we consider a mini-batch SG implementation with momentum in the
Python deep learning library, {\sl Keras} \citep{FC15c}.
To have a fair comparison between SG and subsampled Newton methods, the
following settings are the same for both approaches.
\begin{itemize}
  \item Initial weights.
  \item Network structures.
  \item Objective function.
  \item Regularization parameter.
\end{itemize}
The mini-batch size is $128$ for all SG experiments.
The initial learning rate is selected from $\{ 0.003, 0.001, 0.0003, 0.0001 \}$ by
five-fold cross validation.\footnote{We use a stratified split of data in the
	cross validation procedure.}
  The learning rate is decayed by using {\sl Keras}' default factor $10^{-6}$
and the momentum coefficient is set to be $0.9$. We terminate the training
process after $1,000$ epochs.

The results are shown in Table \ref{table:exps-results-CNN}. For some
sets, the performance of the SG method by using more layers
is inferior to that by fewer layers. It seems overfitting occurs,
so a tuning on SG's  termination criterion may be needed.
Overall, we see that under the same initial settings, the test
accuracy of the subsampled Newton method (with the $5\%$ sampling rate) is
comparable to that of SG.

One question is how long the Newton method takes in comparison with SG. Our
preliminary finding is that Newton is slower but not significant slower.
However, in our settings, a single optimization problem is solved by the Newton
method, while for SG, a cross validation procedure is conducted to select the
initial learning rate. The selection procedure is known to be essential because
of SG's high sensitivity on this parameter. If we take the cross-validation
procedure into consideration, the overall cost of SG is higher. A thorough
timing comparison is much needed and we leave it as a future research issue.

\begin{table}[t]
\caption{Test accuracy by Newton and SG methods. We run $100$ Newton iterations,
while for SG, we run $1,000$ epochs.}
\label{table:exps-results-CNN}
\begin{center}
\begin{tabular}{l >{\centering\arraybackslash}p{3cm} >{\centering\arraybackslash}p{2cm} >{\centering\arraybackslash}p{2cm} >{\centering\arraybackslash}p{2cm}}
\multicolumn{1}{c}{ } & \multicolumn{2}{c}{3-layer CNN} & \multicolumn{2}{c}{5-layer CNN} \\
\hline
	& Newton & SG & Newton & SG\\
\hline
{\sf MNIST} & $99.28\%$ & $99.22\%$ & $99.45\%$ & $99.40\%$ \\
\hline
{\sf SVHN} & $92.72\%$ & $93.00\%$ & $94.14\%$  & $94.46\%$  \\
\hline
{\sf CIFAR10} & $78.52\%$ & $80.17\%$ & $79.72\%$  & $79.65\%$ \\
\hline
{\sf smallNORB} & $95.01\%$ & $95.29\%$ & $95.55\%$ & $93.99\%$
\end{tabular}
\end{center}
\end{table}

\section{Conclusions}
\label{sec:conclu}
In this study, we establish all the building blocks of Newton methods for CNN.
A simple and effective {\sl MATLAB} implementation is developed for public use.
Based on our results, it is possible to further enhance Newton methods for CNN.

\section*{Acknowledgments}
This work was supported by MOST of Taiwan via the grant 105-2218-E-002-033.

\appendix

\section{List of Symbols}
\label{subsec:table-notation}
\begin{longtable}{|p{0.15\textwidth}|p{0.85\textwidth}|}
\hline
Notation & Description\\
\hline
\endhead
\hline
\endfoot
$\by^i$ & The label vector of the $i$th training instance.\\
\hline
$Z^{1,i}$ & The input image of the $i$th training instance.\\
\hline
$l$ & The number of training instances.\\
\hline
$K$ & The number of classes.\\
\hline
$\btheta$ & The model vector (weights and biases) of the neural network.\\
\hline
$\xi$ & The loss function.\\
\hline
$\xi_i$ & The training loss of the $i$th instance.\\
\hline
$f$ & The objective function.\\
\hline
$C$ & The regularization parameter.\\
\hline
$L$ & The number of layers of the neural network.\\
\hline
$L^c$ & The number of convolutional layers of the neural network.\\
\hline
$L^f$ & The number of fully-connected layers of the neural network.\\
\hline
$n_m$ & The number of neurons in the $m$th layer ($L^c < m \leq L+1$).\\
\hline
$n$ & The total number of weights and biases.\\
\hline
$a^m$ & Height of the input image at the $m$th layer ($1 \leq m \leq L^c$).\\
\hline
$a^m_{\padding}$ & Height of the image after padding at the $m$th layer ($1 \leq
m \leq L^c$).\\
\hline
$a^m_{\conv}$ & Height of the image after convolution at the $m$th layer ($1
\leq m \leq L^c$).\\
\hline
$b^m$ & Width of the input image at the $m$th layer ($1 \leq m \leq L^c$).\\
\hline
$b^m_{\padding}$ & Width of the image after padding at the $m$th layer ($1 \leq
m \leq L^c$).\\
\hline
$b^m_{\conv}$ & Width of the image after convolution the $m$th layer ($1 \leq m
\leq L^c$).\\
\hline
$d^m$ & the depth (or the number of channels) of the data at the $m$th layer ($1
\leq m \leq L^c$).\\
\hline
$h^m$ & the height (width) of the filters at the $m$th layer.\\
\hline
$W^m$ & The weight matrix in the $m$th layer.\\
\hline
$\bb^m$ & The bias vector in the $m$th layer.\\
\hline
$S^{m,i}$ & The result of $(W^m)^T \phi(\padding(Z^{m, i})) + \bb^m
\mathds{1}^T_{a^m b^m}$ in the $m$th layer for the $i$th instance ($1 \leq m \leq
L^c$).\\
\hline
$Z^{m+1,i}$ & The output matrix (element-wise application of the activation
function on $S^{m,i}$) in the $m$th layer for the $i$th instance ($1 \leq m \leq
L^c$).\\
\hline
$\bs^{m,i}$ & The result of $(W^m)^T \bz^{m, i} + \bb^m$
in the $m$th layer for the $i$th instance ($L^c < m \leq L$).\\
\hline
$\bz^{m+1,i}$ & The output vector (element-wise application of the activation
function on $\bs^{m,i}$) in the $m$th layer for the $i$th instance ($L^c \leq m
\leq L$).\\
\hline
$\sigma$ & The activation function.\\
\hline
$J^i$ & The Jacobian matrix of $\bz^{L+1, i}$ with respect to $\btheta$.\\
\hline
$\mathcal{I}$ & An identity matrix. \\
\hline
$\alpha_k$ & A step size at the $k$th iteration. \\
\hline
$\rho_k$ & The ratio between the actual function reduction and the predicted reduction at the $k$th iteration. \\
\hline
$\lambda_k$ & A parameter in the Levenberg-Marquardt method. \\
\hline
\end{longtable}

\bibliographystyle{apalike}
\bibliography{sdp}
\end{document}